\documentclass[3p]{elsarticle}

\usepackage{bm}
\usepackage{url}
\usepackage{subfigure}
\usepackage{mathtools}
\usepackage{amssymb}
\usepackage{amsmath}
\usepackage{xcolor}
\usepackage{tikz-cd}
\usepackage{graphicx}
\usepackage{epstopdf}
\usepackage{amsbsy}
\usepackage{doi}
\usepackage[toc,page]{appendix}

\newdefinition{mydef}{Definition}

\newcommand{\Img}{\mathcal{M}}

\newcommand{\Avg}{A}
\newcommand{\Transient}{\mathcal{T}}
\newcommand{\Obs}{f}

\newcommand{\real}{\mathbb{R}}
\newcommand{\cmplx}{\mathbb{C}}

\newcommand{\num}{\mathbb{N}}
\newcommand{\Stat}{\mathcal{S}}
\DeclareMathOperator{\Mes}{Mes}

\DeclareMathOperator{\Prob}{Prob}
\DeclareMathOperator{\Prb}{Prob}
\DeclareMathOperator{\Dens}{Prob}

\DeclareMathOperator{\diag}{diag}
\newcommand{\Linear}{\mathcal{L}}
\newcommand{\DataMes}{\alpha}

\newcommand{\Y}{Y}
\newcommand{\Dq}[1]{\boldsymbol{q}}
\newcommand{\DG}[1]{\boldsymbol{G}}
\newcommand{\DtildeG}[1]{\tilde{\boldsymbol{G}}}
\newcommand{\Dv}[1]{\boldsymbol{v}}
\newcommand{\DV}[1]{\boldsymbol{V}}
\newcommand{\DaH}[1]{\boldsymbol{H}}
\newcommand{\DtildeH}[1]{\tilde{\boldsymbol{H}}}
\newcommand{\longeq}{\scalebox{4}[1]{=}}

\newtheorem{theorem}{Theorem}
\newtheorem{Assumption}{Assumption}
\newtheorem{lemma}[theorem]{Lemma}

\makeatletter
\def\ps@pprintTitle{%
 \let\@oddhead\@empty
 \let\@evenhead\@empty
 \def\@oddfoot{\centerline{\thepage}}%
 \let\@evenfoot\@oddfoot}
\makeatother

\biboptions{sort&compress}

\begin{document}

\title{An information-geometric approach to feature extraction and moment reconstruction in dynamical systems}

\author[courant]{Suddhasattwa Das}
\ead{dass@cims.nyu.edu}

\author[courant]{Dimitrios Giannakis}
\ead{dimitris@cims.nyu.edu}

\author[SDSC]{Enik\H{o} Sz\'ekely\corref{cor}}
\ead{eniko.szekely@epfl.ch}

\cortext[cor]{Corresponding author}

\address[courant]{Courant Institute of Mathematical Sciences, New York University, New York, NY 10012, USA}
\address[SDSC]{Swiss Data Science Center, ETH Z\"urich and EPFL, 1015 Lausanne, Switzerland}

\begin{abstract}

We propose a dimension reduction framework for feature extraction and moment reconstruction in dynamical systems that operates on spaces of probability measures induced by observables of the system rather than directly in the original data space of the observables themselves as in more conventional methods. Our approach is based on the fact that  orbits of a dynamical system induce probability measures over the measurable space defined by (partial) observations of the system. We equip the space of these probability measures with a divergence, i.e., a distance between probability distributions, and use this divergence to define a kernel integral operator. The eigenfunctions of this operator create an orthonormal basis of functions that capture different timescales of the dynamical system. One of our main results shows that the evolution of the moments of the dynamics-dependent probability measures can be related to a time-averaging operator on the original dynamical system. Using this result, we show that the moments can be expanded in the eigenfunction basis, thus opening up the avenue for nonparametric forecasting of the moments. If the collection of probability measures is itself a manifold, we can in addition equip the statistical manifold with the Riemannian metric and use techniques from information geometry. We present applications to ergodic dynamical systems on the 2-torus and the Lorenz 63 system, and show on a real-world example that a small number of eigenvectors is sufficient to reconstruct the moments (here the first four moments) of an atmospheric time series, i.e., the realtime multivariate Madden-Julian oscillation index. 
\end{abstract}

\maketitle

\section{Introduction}

Extracting temporal patterns from data generated by complex dynamical systems is an important problem in modern science with applications in virtually every scientific and engineering domain. The datasets acquired from such systems are increasingly large both in sample size and dimensionality, and it is worthwhile exploring the multitude of data analysis techniques available in the machine learning literature to analyze them. However, many machine learning techniques consider the data points to be independent and identically distributed, and do not take into account the temporal information (i.e., the time ordering of the data), which is a direct outcome of the dynamical evolution taking place in the system's state space. The feature extraction and moment reconstruction method presented here takes into account this information about the dynamics, therefore placing the current work at the intersection of three different fields, namely machine learning, dynamical systems theory, and information geometry. 

In this paper, we take a data-driven approach to the study of dynamical systems. Data-driven methods perform feature extraction by computing the eigenfunctions of a kernel integral operator, i.e., a covariance \citep{PackardEtAl80, BroomheadKing86, AubryEtAl91, GhilEtAl02} or heat \citep{GiannakisMajda12a, BerryEtAl13} operator, acting on functions defined on the state space (observables). These methods often incorporate information about the dynamics by embedding the data in delay-coordinate spaces (Takens time-lagged embedding) \citep{Takens81, SauerEtAl91, Robinson05, DeyleSugihara11}. An alternative approach that has been successfully applied to the analysis of nonlinear dynamical systems is to compute the eigenfunctions of groups or semigroups of operators, e.g., the Koopman or Perron-Frobenius operators \citep{BudisicEtAl12,EisnerEtAl15}, governing the time evolution of observables under the dynamics \citep{DellnitzJunge99, MezicBanaszuk04, Mezic05, GiannakisEtAl15, Giannakis19, BruntonEtAl16, DasGiannakis19, ArbabiMezic16}. Recently, it was established that these two families of techniques can yield equivalent results in an asymptotic limit of infinitely many delays \cite{Giannakis19, DasGiannakis19}. While the above methods are applied directly in the ambient data space, in this paper we propose to work in spaces of probability measures induced by observables of the dynamical system rather than directly in the data space of these observables. These probability measures form together a collection or an ensemble of probability measures that we denote by $\mathcal{S}$. As we will discuss in Sect.\ \ref{subsect:statisticalManifolds}, under certain assumptions the collection $\mathcal{S}$ forms a statistical manifold, i.e., a differentiable manifold whose points are probability measures. This allows us to make connections between the framework presented here and the field of information geometry that studies statistical manifolds.

In recent years there has been an increased interest in extracting dynamical patterns by working in probability spaces \citep{MuskulusVerduyn11, TalmonCoifman13, LianEtAl15, DsilvaEtAl16} instead of the ambient data space. In particular, \cite{TalmonCoifman13} and \cite{LianEtAl15} consider that the observations are drawn from conditional time-varying probability density functions (PDFs). Representative (temporal) patterns on the manifold can be extracted by employing feature extraction techniques, such as the Diffusion Maps algorithm \cite{CoifmanLafon06} with a kernel based on statistical distances, i.e., divergences, between the PDFs. Another approach \cite{MuskulusVerduyn11} is to embed multiple dynamical systems into a single low-dimensional space by comparing their invariant measures. In this approach, phase space trajectories are interpreted as probability measures, and a distance-based embedding method, namely multidimensional scaling \cite[MDS;][]{CoxCox94}, is applied to the Wasserstein distance matrix to perform feature extraction and uncover the low-dimensional manifold. Information-geometric techniques have also been employed in data analysis applications such as flow cytometry or document classification \citep{CarterEtAl09, CarterEtAl11}. In these cases, the PDFs are estimated over subsets of sample populations, and the distances between (similar enough) PDFs on the statistical manifold are computed using approximations of the Fisher information distance, such as the Kullback-Leibler divergence, the Hellinger distance or the cosine distance. Subsequently, the distances between all pairs of PDFs are approximated by geodesic distances on the statistical manifold, and the full distance matrix is then embedded into a low-dimensional manifold using MDS techniques.

In this paper, we consider to have access to (potentially partial) observations of a deterministic ergodic dynamical system through some scalar or vector-valued observable $f$. Finite-time trajectories along the dynamical system induce probability measures over the measurable space defined by the observation map $f$. One of our standing assumptions is that the data is drawn from observations on an ergodic trajectory. We will make our assumptions more precise later, but intuitively, ergodicity ensures that time averages of a quantity on a typical trajectory converge to its space average, as one takes longer and longer averaging windows. Thus an ergodic trajectory fills out the ambient space with a distribution that is the same as that of an invariant measure of the dynamics. We equip the probability space defined by the collection of probability measures with a divergence, here the Hellinger distance, and use it to define a symmetric and positive definite kernel over the space of measures. We demonstrate that the eigenfunctions of a kernel integral operator computed using the Diffusion Maps algorithm \citep{CoifmanLafon06, TalmonCoifman13, LianEtAl15} capture temporal and spatiotemporal patterns of interest of the dynamical system, and are meaningful for dimension reduction and moment reconstruction. Under suitable assumptions, the probability measures described above lie on a statistical manifold equipped with a Riemannian metric, namely the Fisher information metric. The distance induced by this metric, i.e., the Fisher information distance, can be approximated by the Hellinger distance, thus further justifying our choice of divergence in the probability space.

An important contribution that we propose in this paper is a novel method to reconstruct the $n$-th moment of the probability measures induced by trajectories along the dynamical system, for every $n\in\mathbb{N}$. The moment reconstruction is done in the Hilbert space defined by the basis of eigenfunctions of a kernel integral operator constructed in the space of probability measures. One of our key results is an identity (Theorem \ref{thm:A}) that expresses these moments as time-averaging operators of a certain function on the collection of probability measures. We expand this result into a more general identity (Theorem \ref{thm:B}) that applies to any continuous function on the data space, not just moments. We demonstrate the reconstruction of moments on an atmospheric time series index (the realtime multivariate Madden-Jullian oscilation index), where we show that a small number of leading eigenvectors is sufficient to accurately recover the moments. This opens up the avenue for nonparametric forecasting of the moments of the distributions, and therefore of the distributions themselves. 

\section{Contributions and outline of the paper}\label{sect:outline}

The main goal of the paper is to present a data-driven technique for the reconstruction of finite-time statistics, e.g., moments, of probability measures defined on trajectories of a dynamical system. To enable that, we lay down a rigorous theoretical framework and prove that the data-driven vectors and matrices that we use in our numerical methods converge to functions and operators on associated underlying spaces. In order to do that, we introduce several spaces and mappings over the course of the paper, and we list the most important ones in Table \ref{tab:spaces}. 

\setlength{\arrayrulewidth}{0.2mm}
\begin{table}[]
\begin{tabular}{ |p{0.105\textwidth}|p{0.85\textwidth}|  }
\hline
\textbf{Notation} & \textbf{Description}  \\
\hline
\hline
$X$ & state space of the dynamical system \\ \hline
$\Mes(\Y)$ & Borel complex valued measures on $\Y$  \\ \hline
$\Prob(\Y)$ & Borel probability measures on $\Y$\\ \hline
$\Prob_c(\Y)$ & Borel probability measures on $\Y$ with compact support \\ \hline
$\Dens(\Y;\DataMes)$ & Borel probability measures on $\Y$ which are absolutely continuous wrt a reference measure $\DataMes$\\ \hline
$\Stat$  & Collection of probability measures under the map $p: X \to \Prob_c(\Y)$ \\ 
\hline
\hline
\textbf{Inclusions}: & $\Stat \subset \Prob_c(\Y) \subset \Prob(\Y) \subset \Mes(Y)$ \\ 
 & $\Dens(\Y;\DataMes) \subset \Prob(\Y)$\\
\hline 
\end{tabular}
\caption{Summary of the spaces used in the paper.}
\label{tab:spaces}
\end{table}

Assumption~\ref{asmptn:1} in Sect.~\ref{sect:statistics} introduces the basic assumptions that we impose on the system, and Assumption~\ref{asmptn:data} in Sect.~\ref{sect:approx} introduces the assumption made on the data available. One of our main contributions is Theorem~\ref{thm:A} where we show that the moments of probability measures induced by finite-time trajectories of a dynamical system can be obtained by averaging an associated observation map over the respective trajectories. In Sect.~\ref{sect:statistics} we give a more general version (Theorem~\ref{thm:B}) of Theorem~\ref{thm:A}. Theorem~\ref{thm:B} is stated for integrals of arbitrary continuous functions with respect to a collection of probability measures. The various moments of a probability measure discussed in Theorem~\ref{thm:A} are a special case of such integrals. 

Section~\ref{sect:geometry} describes several geometric aspects, e.g., embeddings of probability measures and invariance under isometries of the data, using a divergence between probability measures, i.e., the Hellinger distance. If the collection $\Stat$ is a manifold, it can be equipped with a Riemannian metric and the natural divergence associated with it can be approximated by the Hellinger distance, thus allowing us to make connections to the field of information geometry.  

The data-driven aspect of our framework and numerical approximations of the continuous quantities introduced in the theorems is addressed in Sect.~\ref{sect:approx}. Algorithm 1 shows the feature extraction in the probability space, and Algorithm 2 shows the moment reconstruction using a small number of leading eigenvectors. Theorem~\ref{thm:data} establishes the almost sure convergence of our numerical methods. The proof of all the theorems is done in Sect.~\ref{sect:proofs}. In Sect.~\ref{sect:exp} we study three low-dimensional dynamical systems and a real-world atmospheric time series index, namely the realtime multivariate Madden-Julian oscillation (RMM) index. We end with some concluding remarks and future perspectives in Sect.\ \ref{sect:conclusion}.

\section{Dynamics-dependent probability measures on the data space}\label{sect:statistics}

The following will be a standing assumption in the rest of our discussions.

\begin{Assumption}\label{asmptn:1}
$\hat{X}$ is a $C^1$-manifold (differentiable manifold) with a $C^1$ deterministic flow $\Psi_t :\hat{X} \mapsto \hat{X}$. There exists a $\Psi_t$-invariant ergodic, probability measure $\mu$ for the flow, with a compact support $X\subset \hat{X}$. $X$ is equipped with its Borel $ \sigma $-algebra $ \mathcal{ B}(X) $. $f:\hat{X} \mapsto \Y $ is a $C^1$ observation map taking values on a manifold $\Y$. 
\end{Assumption}

In the examples from Sects.~\ref{sect:2torus} and~\ref{sect:Oxtoby} ahead, $ X = \hat{X} = \mathbb{T}^2$, the 2-torus, whereas in Sect.~\ref{sect:lorenz}, $\hat{X}=\real^3$ and $X$ is the Lorenz 63 strange attractor. The space $\Y$ plays the role of a data space, and is often the Euclidean space $\real^d$. 
The triple $(X, \Psi_t, \mu)$ defines a measure-preserving dynamical system. One can associate to each state $x \in X$ a trajectory $\{\Psi_t(x)\}_{t\in\real}$. Our focus will be on the collection of probability measures induced by the observation map $f$ on finite-time trajectories at each point, and their associated statistics, such as the moments of these probability measures.  

\subsection{Probability measures on the data space} 

Let $ T = [ - \Delta t, 0 ] $, $ \Delta t > 0 $,  be a closed time interval, $\mathcal{ B }(T)$ be its Borel $\sigma$-algebra, and $ \lambda$ be the  Lebesgue probability measure on $T$. Next define a map $g$  which assigns to every state $x \in X$ a continuous, $\Y$-valued map $g_x = g(x)$ on $T$. $g_x$ is defined as $g_{x}(t) = f\left( \Psi_t( x ) \right)$ for every $t\in T$. Thus $g$ is a mapping from $X$ to the set of continuous functions defined on $T$ with values in $Y$, $g : X \mapsto C^0( T; \Y)$. Let $\Prb(\Y)$ denote the set of Borel probability measures on $\Y$, and $\Prb_c(Y)$ denote its subset of compactly supported measures. Let $ \Lambda : C^0( T; \Y) \mapsto \Prb_c(\Y)$ be the map defined as $\Lambda : h \mapsto h_{*}\lambda$. Here $h_*\lambda$ is the \emph{push-forward} of the Lebesgue probability measure $\lambda$ on $T$ to a Borel probability measure on $\Y$, defined as $h_*\lambda(A)$ = $\lambda(h^{-1}(A))$ for every Borel set $A\subset\Y$.  Now define the map $p:X\to \Prb_c(\Y)$ as 
\begin{equation}\label{eq:probMeasure} 
p:= \Lambda\circ g; \quad p_x = p(x) = g_{x*}\lambda, \quad p_x(A) = \lambda \left\{ t\in T : f(\Psi_t(x))\in A \right\}, \quad \forall A\in\mathcal{B}(\Y), \quad \forall x\in X.  
\end{equation}

The image of the observation map $f$ will be denoted as $\Img \subset \Y$, and the image of $p$, namely the set $p(X) = \{ p_x \mid x \in X \}$, will be denoted as $\Stat$. Thus $\Stat \subset \Prb(\Img) \subset \Prb_c(\Y) \subset \Prb(\Y)$. Our main focus in this paper will be this collection $ \Stat$ of probability measures $p_x$. They represent the set of all possible probability measures obtained from trajectories of length $\Delta t$ from initial states in $X$. 

In a typical data-driven setting, $X$ and its dynamics are unknown, and we only observe data points $y=f(x)$, $y \in Y$, as functions of the unobserved states $x$. In the operator-theoretic framework, e.g., Koopman or Perron-Frobenius operators, one studies the effect of the dynamics on the space of observables on $X$, i.e., the data space $\Y$. In the framework presented here, we go one step further from a typical data-driven setting and consider observables $\gamma$ on the data space $Y$. These indirectly lead to observables on $X$ through composition with $f$. Namely, any function $\gamma:\Y\to \real^m$ induces the function $\gamma\circ f: X\to \real^m$.  Here we consider the set of probability measures $\Prb(\Img)$ or $\Prb_c(\Y)$. As we will see in the following, the moments of the probability measures $p_x$ are a special case of the observables $\gamma$, and are of particular interest in this paper.

\subsection{Moments of the probability measures in $\mathcal{S}$} 

Consider the case when $Y=\real^d$. Every $p_x$ in the collection $\Stat$ is a probability measure with a compact (bounded) support in the set $\Img \subset \Y = \real^d$. Therefore all of its moments exist and are finite. Let $\gamma^{(n)} : \real^{d} \to \real^{d}$ be the function that raises each component of a vector to the power $n$, i.e.,
\begin{equation}\label{eq:defGammaMoments}
\gamma^{(n)} : \real^{d} \to \real^{d}; \quad (y^1, \ldots, y^d) \mapsto ((y^1)^n, \ldots,(y^d)^n).
\end{equation}

Then for every $p_x \in \Stat$ and $n \in \mathbb{N}$, the $d$-dimensional vector of the $n$-th moment $\mathbb{E}_n$ of $p_x$ is defined as
\[
\mathbb{E}_n :  \Stat\to \real^{d}; \quad \mathbb{E}_n: p_x \mapsto \int_{\Y} \gamma^{(n)}(y) d p_x(y)dy.
\]
Therefore, $\mathbb{E}_n(p_x)$ is the $d$-dimensional vector whose $i$-th coordinate $\mathbb{E}_n^{(i)}(p_x) = \int_{ \Y } (y_i)^n d p_x(y)dy$. Our first main result (Theorem \ref{thm:A} below) expresses these moments as a function of the initial state $x\in X$, in terms of the following time-averaging operator $\Avg_{\Delta t}$:
\[ \Avg_{\Delta t} : L^2(\mu) \to L^2(\mu), \quad \Avg_{\Delta t} f :  x\mapsto \frac{1}{\Delta t} \int_{-\Delta t}^0 f (\Psi_t( x))dt. \]

\begin{theorem}\label{thm:A}
Let Assumption \ref{asmptn:1} hold and $\Y=\real^d$. Then for every $n\in\mathbb{N}$ and $x\in X$, the n-th moment $\mathbb{E}_n(p_x)$ is finite and 
\[\mathbb{E}_n(p_x) = \Avg_{\Delta t} (\gamma^{(n)}\circ f ) (x) .\]
\end{theorem}
Taking for example $n = 1$, we have $\gamma^{(n)}=\gamma^{(1)} = \mathbb{I}_{\Y}$, the identity map on $\Y$, and the first moment (mean) $\mathbb{E}_1$ is a time-averaging operator acting on the observable $f$ since $\Avg_{\Delta t}(\gamma^{(1)} \circ f)=\Avg_{\Delta t}(f)$. 

\paragraph{Remark} A (probability) measure is completely characterized by its moments, and Theorem \ref{thm:A} expresses these moments in terms of a time-averaging operator. Note that in a data-driven setting, the observation map $f$ is fixed and its values are known at some sampling points $x_i$. The function $\gamma^{(n)} \circ f$ can be computed to any desired degree of accuracy at these sampled points. The operator $\Avg_{\Delta t}$ involves an integral which can be numerically approximated by averaging along a sampled trajectory. We later show in \eqref{eqn:change_var} that Theorem \ref{thm:A} is a special case of an identity that involves more general functions $\gamma:\Y\to \real^m$, e.g., Fourier functions, spherical harmonics or polynomials, not just $\gamma = \gamma^{(n)}$ from \eqref{eq:defGammaMoments}.

\subsection{Observables on the data space} 

Recall that the space $\Prb_c(\Y)$ is a convex subset of the linear space $\Mes(\Img)$ of finite, complex measures on $\Img$. Let $\Linear( \Mes(\Img) ; \real^m )$  denote the set of linear maps from $\Mes(\Img)$ into $\real^m$ which are bounded (and thus continuous). Now define the map
\begin{equation}\label{eq:linTransf}
J : C^0(\Img;\real^m) \to \Linear( \Mes(\Img) ; \real^m ); \quad J(\gamma) : \pi \mapsto \int_{\Y} \gamma d\pi, \quad \forall \pi \in \Mes (\Img).
\end{equation}
Since for every function $\gamma \in C^0(\Img;\real^m)$, $J(\gamma)$ is a linear map on $\Mes (\Img)$, it automatically becomes a continuous map on $\Stat$, which is a collection of probability measures contained in $\Mes (\Img)$. Moreover, for every $ a_1, a_2 \in \cmplx$ and every $\gamma_1, \gamma_2 \in  C^0(\Img;\real^m)$,
\[J(a_1 \gamma_1 + a_2 \gamma_2)(\pi) = a_1 \int_Y \gamma_1 d \pi + a_2 \int_Y \gamma_2 d \pi =  a_1  J(\gamma_1)(\pi) + a_2 J(\gamma_2)(\pi),\]
making $J$ a linear map. Another way to view the action of $J$ is as a dual mapping. The space $\Mes(\Img)$ lies in the dual space to the Banach space $C^0(\Img;\cmplx)$, and $C^0(\Img;\cmplx)$ embeds isomorphically and canonically into the double dual $C^0(\Img;\cmplx)^{**} = \Mes(\Img)^{*}$. Then it follows from (\ref{eq:linTransf}) that, by definition, $J$ maps each component of $\gamma$ into its double dual. The following important theorem expresses the action of $J(\gamma)$ on elements of $\Stat$, $p_x \in \Stat$, in terms of the time-averaging operator $\Avg_{\Delta t}$, for every $\Delta t>0$.

\begin{theorem}\label{thm:B}
For every $\gamma \in C^0(\Img;\real^m)$ and every $x\in X$,
\begin{equation}\label{eqn:change_var}
J(\gamma)(p_x) = \frac{1}{\Delta t} \int_{-\Delta t}^0 \left(\gamma \circ \Obs(\Psi_t (x))\right)dt  = \Avg_{\Delta t}\left( \gamma\circ\Obs \right)(x).
\end{equation}
\end{theorem}

The diagram below illustrates the domains and codomains of the various maps defined so far, and how they are connected through \eqref{eqn:change_var}. 
\[\begin{tikzcd}[]
X \arrow{r}{f} \arrow{d}{p_x}
& \Img \arrow{r}{\subset}
&\Y \arrow{r}{\gamma}
& \real^{m} \arrow[dashed]{d}{A_{\Delta t}}
\\ \Stat \arrow{r}{\subset}
& \Prb (\Img) \arrow{r}{\subset}
& \Prb_c (\Y) \arrow{r}{J(\gamma)}
&  \real^{m}
\end{tikzcd}\]

We will next discuss various geometric aspects of the collection $\Stat$ of probability measures.

\section{Geometrical structures on the collection of probability measures}\label{sect:geometry}

In the following we define a  metric on the space $\Prb_c(\Y)$ of compactly supported Borel probability measures on $\Y$ in a way such that the map $p : X\to \Prb_c(\Y)$ is continuous. One initial difficulty in defining such a metric is that for different $x, x'\in X$, the measures $p_x, p_{x'}$ are highly singular measures with respect to (wrt) the Lebesgue measure on $\Y$. Moreover, their supports are one-dimensional non-intersecting curves. Thus the task is to define a concept of distance between measures supported on almost disjoint curves lying in high-dimensional ambient data spaces. This can be achieved by smoothing out these singular measures using a $C^r$ kernel function $k : \Y \times \Y \to \real^+_0$, by a procedure called \emph{kernel density estimation} \cite{Bowman1997}. Let $\Mes_c(\Y)$ denote the space of finite signed measures on $\Y$ with compact support, and $C^r(\Y)$ denote the set of functions $r$ times differentiable on $\Y$. Then one has the following map
\begin{equation}
K : \Mes_c(\Y) \to C^r(\Y); \quad K(\pi) = \int_{\Y} k(\cdot, y) d \pi(y), \quad \forall \pi \in \Mes_c(\Y).
\end{equation}
It is easy to check that when restricted to $\Prob_c(\Y)$, we have a map $K : \Prob_c(\Y) \to C^r(\Y; \real^+_0)$ . Now let $\DataMes$ be any reference measure on $\Y$. Given a $C^0$ (continuous) non-negative function $h : \Y\to \real^+_0$, $h \DataMes$ is a new measure which is absolutely continuous wrt $\DataMes$ and with a $C^r$ density $h$. Let $\Dens(\Y; \DataMes)$ denote the set of Borel measures on $\Y$ which are $\DataMes$-a.c.\ (i.e., absolutely continuous wrt $\DataMes$), and with $C^0$ density functions. Thus we have the canonical embedding
 \[ \iota_{\DataMes} : C^0(\Y; \real^+_0) \to \Dens(\Y; \DataMes), \quad  \iota_{\DataMes} (h) := h\DataMes. \]

We summarize the various maps we have described in the commutative diagram\footnote{A commutative diagram is a graph comprised of vertices and directed edges. The vertices correspond to some sets, and edges represent maps between these sets. Any path between two points thus correspond to a composition of maps. Commutative diagrams schematically lay out how the various sets are related to each other through maps. The other important information contained in commutative diagrams are the \emph{commuting relations}. If two points are connected by two different paths, then the two maps they represent must be equal. Thus commutative diagrams also summarize the various identities between maps.} below.
\[\begin{tikzcd}[column sep = large]
X \arrow{d}{g} \arrow{dr}{p} \arrow[dashed]{r}{p^{\DataMes}} &\Dens(\Y; \DataMes) &C^0(\Y; \real^+_0)  \arrow{l}[swap]{\iota_{\DataMes}}\\
C^0(T; \Y) \arrow{r}{\Lambda} &\Prob_c(\Y) \arrow{r}{K} & C^r(\Y; \real^+_0) \arrow{u}{\subset} 
\end{tikzcd}\]
The dashed arrow defines the map 
\[p^{\DataMes} : X\to \Dens(\Y; \DataMes); \quad p^{\DataMes} = \iota_{\DataMes} \circ K \circ p; \quad p^{\DataMes}_x = p^{\DataMes}(x) = \iota_{\DataMes} K (p_x), \]
which embeds the state (phase) space $X$ of the dynamics into a space of measures which are absolutely continuous wrt $\DataMes$. 

\paragraph{Choice of kernel} The kernel $k$ is usually taken to be an isotropic kernel, i.e., of the form 
\begin{equation}\label{eqn:def:isotropic}
	k(y,y') = \eta\left( d(y,y') \right), \quad \forall y, y' \in Y,
\end{equation}
where $\eta : \real\to \real^+_0$ is the so-called \emph{shape} function, and $d(y, y')$ is the distance between data points $y$ and $y'$, e.g., the Euclidean distance. We are going to need the following assumption on the kernel : 

\begin{Assumption}\label{asmptn:kernel}
$k:\Y\times\Y\to\real_0^+$ is a $C^r$, strictly positive definite, isotropic kernel as in \eqref{eqn:def:isotropic}. Further, its shape function $\eta$ satisfies $\lim_{|d|\to\infty} \eta(d) = 0$, with $d$ the distance measure.
\end{Assumption}

\subsection{The Hellinger distance}\label{subsect:Hellinger}

In order to study the geometrical properties of the collection of probability measures $\mathcal{S}$ we need to equip the probability space with a divergence, i.e., a notion of dissimilarity between probability measures. We choose the Hellinger distance over other divergences (e.g., Kullback Leibler divergence (relative entropy), Wasserstein distance, total variation distance) due to several reasons that will become clearer in Sects.~\ref{subsect:Hellinger}--\ref{subsect:statisticalManifolds}.

Given two probability distributions $\pi$ and $\pi'$ which are absolutely continuous wrt the probability measure $\DataMes$ with densities $\rho, \rho'$, the squared Hellinger distance is defined as
\begin{equation}\label{eq:Hellinger}
	d_H^2(\pi,\pi') = \int_{\real^d} \left(\sqrt{\rho} - \sqrt{\rho'}\right)^2 d \DataMes.
\end{equation}
This definition is usually accompanied by a multiplicative factor of $1/2$, which we have dropped. We will now establish some conditions under which $p^{\alpha}$ is an embedding into $\Prb_c(Y)$.

{\begin{theorem}\label{thm:D}
Let Assumption \ref{asmptn:1} hold, then for every $x\in X$, $p_x$ has a compact support contained in $\Img$, and $p:X\to \Prb(\Img)$ is a continuous map in the weak topology on $\Prb_c(\Y)$ (and $\Prb_c(\Img)$). Further, let Assumption \ref{asmptn:kernel} hold. Then $p^{\DataMes} : X \to \Dens(\Y; \DataMes)$ is a continuous map wrt the Hellinger distance. If in addition, the observation map $f:X\to Y$ is one-to-one, then $p$ is a homeomorphism between $X$ and $\Stat$, and $p^{\DataMes}$ is injective.
\end{theorem}

\paragraph{Remark} If $X$ is a manifold and $p$ is injective, then $\Stat$ can be assigned the same manifold structure as $X$, and $\Stat$ becomes a \emph{statistical manifold} of probability measures, parameterized by the manifold $X$. With this manifold property, techniques from the field of information geometry \citep{AmariNagaoka07} (Sect.\ \ref{subsect:statisticalManifolds}) can be employed to design data analysis algorithms taking advantage of the geometrical structure of $ \mathcal{ S } $. In Sect.\ \ref{sect:lorenz}, we will demonstrate with numerical experiments on the Lorenz 63 system that these techniques remain useful even if the state space $ X $ and/or $ \Stat$ are not smooth manifolds. Statistical manifolds carry a lot of information intrinsic to the underlying dynamical system and are easily tractable from a data point of view.

Note that the maps $g$, and therefore $p$ and $p^{\DataMes}$, depend on the observation map $f:X \to \Y$. We would like to have invariance of the collection of probability distributions under transformations of the dataset $\Img = f(X)$ such as translation and rotation. We show in the following that this is achieved by our kernel-based embedding of $\Prob_c(\Y)$ into the space $\Prob(\Y; \DataMes)$ of $\DataMes$-a.c.\ measures. 

\paragraph{Diffeomorphisms of the image} Let $D: \Y \to\Y'$ be a $C^r$ map which maps $\Y$ diffeomorphically into its image. Then the observation map $f:X \to \Y$ is transformed into $f' = D\circ f : X \to \Y'$. This new observation map $f'$ results in a new map $g':X \to C^0(T; \Y')$ given by $g'_x(t)$ = $f'(\Psi_t (x))$. One can similarly define a new map $p'=\Lambda' \circ g'$ and a new statistical manifold $\Stat' = \{p'_x\ |\ x \in X\}$, with $ \Lambda' : C^0( T; \Y') \mapsto \Stat'$, $\Stat' \subset \Prb_c(Y')$. The following commutative diagrams succinctly display the relations between these two sets of maps.
\[
\begin{tikzcd}[row sep = large, column sep = large]
\Y  \arrow{d}{D} &\ X \arrow[swap]{l}{f} \arrow{dl}{f'}\\
\Y' & \\
\end{tikzcd}
\hspace{1cm}
\begin{tikzcd}
C^0(T; \Y) \arrow{rr}{D \circ} \arrow{dd}{ \Lambda} &\ & C^0(T; \Y') \arrow{dd}{\Lambda'} \\
\ & X \arrow{dr}{p'} \arrow{dl}{p} \arrow{ur}{g'} \arrow{ul}{g} &\ \\
\Stat \arrow{rr}{D_*} &\ & \Stat'  \\
\end{tikzcd}
\]
The map $D_*:\Stat \to \Stat'$ in the diagram is the push-forward of probability measures on $\Img$ under the map $D$, and the map $D \circ: C^0(T;\Y) \to  C^0(T;\Y') $ is left composition by $D$. 

\begin{theorem}\label{thm:E}
Let Assumption \ref{asmptn:1} hold, and $D : \Y \to \Y'$ be the diffeomorphism as above. Suppose that the kernel functions $k:Y\times Y\to \real$ and $k':Y'\times Y'\to\real$ satisfy
\[ k'( Dy, Dy' ) = k(y, y'), \quad \forall y,y'\in Y . \]
Then we have the following commutative diagram.
\[\begin{tikzcd}[column sep = large]
\Y \arrow{d}{D} & X \arrow[swap]{l}{f} \arrow{dl}{f'} \arrow{r}{g} \arrow{rd}[swap]{g'} \arrow[bend left = 20]{rr}{p} \arrow[bend right = 60]{drr}{p'}  & C^0(T; Y) \arrow{r}{\Lambda} \arrow{d}{D\circ} &\Prob_c(Y) \arrow{r}{K} \arrow{d}{D_*} & C^r(Y; \real^+_0) \arrow{r}{\iota_{\DataMes}} &\Dens(Y; \DataMes) \arrow{d}{D_*}[swap]{\cong} \\
\Y' &\ &C^0(T; Y') \arrow{r}{\Lambda'}  &\Prob_c(Y') \arrow{r}{K'} & C^r(Y'; \real^+_0) \arrow{r}{\iota_{D_*\DataMes}} \arrow{u}{\circ D} &\Dens(Y'; D_*\DataMes)
\end{tikzcd}\]
The rightmost map $D_* : \Dens( \Y; \DataMes) \to \Dens(\Y'; D_* \DataMes)$ is an isomorphism wrt the Hellinger distances of the respective spaces.
\end{theorem}

The theorem is proved in Section~\ref{sect:proof:E}.

\paragraph{Remark} If $k$ is an isotropic kernel \eqref{eqn:def:isotropic}, then Theorem~\ref{thm:E} says that the pull-back metric on $X$, given by
\begin{equation}\label{eqn:def:induced_dist}
d_{\DataMes}^2 (x,x') := d_{H}^2 \left( p^{\DataMes}_x , p^{\DataMes}_{x'} \right) = \int_{\Y} \left[ \sqrt{\frac{d (\iota_{\DataMes} K(p_x))}{d\DataMes}} - \sqrt{\frac{d (\iota_{\DataMes} K(p_{x'}))}{d\DataMes}} \right]^2 d\DataMes = \int_{\Y} \left[ \sqrt{ K(p_x) } - \sqrt{ K(p_{x'}) } \right]^2 d\DataMes 
\end{equation}
is invariant under isometries $D: \Y \to \Y$, i.e., diffeomorphisms which preserve the metric of $\Y$.

\paragraph{The measure $\DataMes$} Note that we did not place any restriction on the choice of the measure $\DataMes$. A natural choice for $\DataMes$ is the push-forward of the invariant measure of the dynamics $\mu$ on $X$, i.e., $\DataMes = f_* \mu $. Thereby, we can take advantage of the ergodicity of $\mu$ and approximate $\DataMes$ by the samples $y_i = f(x_i)$ on the data space. We use here for $\DataMes$ the Lebesgue measure which is consistent with our definition of probability measures in~\eqref{eq:probMeasure}, but the above formula does not depend on the type of measure used as reference, i.e., the Hellinger distance will not change if the densities are defined relative to a different equivalent measure. Moreover, we have the following commutative diagram, similar to Theorem~\ref{thm:E}. 
\[\begin{tikzcd}[column sep = large]
\Y \arrow{d}{D} & X \arrow[swap]{l}{f} \arrow{dl}{f'} \arrow{r}{g} \arrow{rd}[swap]{g'} \arrow[bend left = 20]{rr}{p} \arrow[bend right = 60]{drr}{p'}  & C^0(T; Y) \arrow{r}{\Lambda} \arrow{d}{D\circ} &\Prob_c(Y) \arrow{r}{K} \arrow{d}{D_*} & C^r(Y; \real^+_0) \arrow{r}{\iota_{f_*\mu}} &\Dens(Y; f_*\mu) \arrow{d}{D_*}[swap]{\cong} \\
\Y' &\ &C^0(T; Y') \arrow{r}{\Lambda'}  &\Prob_c(Y') \arrow{r}{K'} & C^r(Y'; \real^+_0) \arrow{r}{\iota_{f'_*\mu}} \arrow{u}{\circ D} &\Dens(Y'; f'_*\mu)
\end{tikzcd}\]

\subsection{Kernel integral operators on $\mathcal{S}$}\label{subsect:eigenfunctions}

Given a choice of a reference measure $\DataMes$ on $\Y$, one gets the distance $d^{\DataMes}$ on the state space $X$ \eqref{eqn:def:induced_dist} induced by the Hellinger distance. Using $d^{\DataMes}$ we will define a new kernel on $\Prob_c(\Y)$ (or $\Stat$), defined similarly to the isotropic kernel \eqref{eqn:def:isotropic}, with the shape function $\eta$ chosen to be the Gaussian function
\begin{equation}\label{eq:def:Gauss}
	\eta(d) = e^{-|d|^2/\epsilon},
\end{equation}
where $\epsilon>0$ is a bandwidth parameter. The resulting Gaussian kernel on $\Prob_c(\Y)$ is given by
\begin{equation} \label{eqn:def:kH} 
k_H(\pi, \pi') := \exp\left( - \frac{1}{\epsilon} d_H^2 \left( \iota_{\DataMes}( K\pi ), \iota_{\DataMes} ( K\pi' ) \right) \right) \stackrel{ \text{by \eqref{eq:Hellinger}} } {\longeq} \exp \left( - \frac{1}{\epsilon}\int_{\Y} \left[ \sqrt{K\pi} - \sqrt{K\pi'} \right]^2 d\DataMes \right).
\end{equation}
In the particular instance when both $\pi, \pi'$ lie on $\Stat$ and equals $p_x$, $p_{x'}$ respectively,
\[ k_H(p_x, p_{x'}) := \exp\left( - \frac{1}{\epsilon} d_H^2 \left( p^{\DataMes}_x, p^{\DataMes}_{x'} \right) \right) = \exp \left( - \frac{1}{\epsilon}\int_{\Y} \left[ \sqrt{K(p_x)} - \sqrt{K(p_{x'})} \right]^2 d\DataMes \right). \]
Since $p$ maps $X$ into $\Stat$, it induces a probability measure $\nu$ on $\Stat$, which is the push-forward of the $\Psi_t$-invariant probability measure $\mu$ on $X$, i.e., $\nu = p_*\mu$ . Thus for any Borel measurable set $U\subseteq \Stat$, $\nu(U) = \mu\{ x\in X\ |\ p_x\in U \}$. As a result, one can define a kernel integral operator $G:L^2(\nu) \to L^2(\nu)$ as
\[ G (\phi)(p_x) := \int_{\Stat} k_H(p_x, \pi) \phi(\pi)d\nu(\pi) = \int_{X} k_H(p_x, p_{x'}) \phi(p_{x'}) d\mu(x'),\]
where the second equality above follows from the change of variables formula for integrals and the definition of $\nu$. Each function $\phi$ in the range of $G$, also a member of $L^2(\nu)$, is a pointwise defined function continuous wrt the weak topology on $\Stat$. Let $1_{\Stat}$ denote the constant function equal to $1$ on $\Stat$. Next, we define a sequence of normalizations as in the Diffusion Maps algorithm \cite{CoifmanLafon06} :
\[ q := G(1_{\Stat}), \quad \tilde{G} (\phi) :=  G\left( \frac{\phi}{ q } \right), \quad v :=  \tilde{G}(1_{\Stat}), \quad H (\phi) := \frac{1}{v} \tilde{G}(\phi), \]
to get a Markov integral operator $H:L^2(\nu) \to L^2(\nu)$. The function $v$ is a continuous function on $\Stat$ and is positive everywhere, therefore it can be interpreted as a density function. The operator $H$ is not symmetric, but if $V$ and $Q$ denote the multiplication operators by the functions $v$ and $q$, respectively, then
\[ \tilde{H} := V^{\frac{1}{2}} H V^{-\frac{1}{2}} = V^{\frac{1}{2}} V^{-1} \tilde{G} V^{-\frac{1}{2}} = V^{-\frac{1}{2}} \tilde{G} V^{-\frac{1}{2}} = V^{-\frac{1}{2}} Q^{-1} G Q^{-1} V^{-\frac{1}{2}} , \]
is a symmetric operator. Moreover, since $v$ is a continuous function, positive and uniformly bounded away from $0$, $V$ and $V^{-1}$ are both bounded operators. Thus, $\tilde{H}$ is also compact and self-adjoint, and it has a complete eigenbasis consisting of functions $\{ \psi_l \}_{l=1}^{\infty} $ with eigenvalues $1 - \lambda_l$:
\begin{equation}
\tilde{H} \psi_l = (1 - \lambda_l) \psi_l.
\end{equation}
The eigenfunctions $\psi_l$ are orthonormal in $L^2(\nu)$, i.e., wrt the inner product and the measure $\nu$: 

\[ \langle \psi_k, \psi_l \rangle_{\nu} = \int_{\Stat} \psi_k(\pi)^* \psi_l(\pi) d{\nu}(\pi) = \delta_{kl} .\]

Since $H$ and $\tilde{H}$ are related by a similarity transformation, $H$ has the same spectrum as $\tilde{H}$. Thus $H$ has the same eigenvalues as $\tilde{H}$ with the corresponding eigenfunctions $\phi_l$ satisfying
\begin{equation}\label{eq:phis}
	H \phi_l = (1 - \lambda_l) \phi_l, \quad l=1, 2, \ldots, \quad \phi_l := V^{-\frac{1}{2}} (\psi_l) = v^{-\frac{1}{2}} \psi_l.
\end{equation}

Let $\omega$ be the measure whose density wrt $\nu$ is $v$, i.e., $d\omega(\pi) = v(\pi) d\nu(\pi)$. Then the $\phi_l$s form a basis, which is not orthonormal in $L^2(\nu)$ but orthonormal in $L^2(\omega)$, namely,
\[ \langle \phi_k, \phi_l \rangle_{\omega} = \int_{\Stat} \phi_k(\pi)^* \phi_l(\pi) d\omega(\pi) = \int_{\Stat} \phi_k(\pi)^* \phi_l(\pi) v(\pi) d\nu(\pi) = \int_{\Stat} \psi_k(\pi)^* \psi_l(\pi) d\nu(\pi) = \langle \psi_k, \psi_l \rangle_{\nu} = \delta_{kl} .\]

The eigenfunctions $\phi_l$ of the Markov matrix $H$ are also the eigenfunctions of the random walk Laplacian operator $L=I-H$, i.e.,
\begin{equation}\label{eq:phisLaplacian}
	L \phi_{l} = \lambda_{l} \phi_{l},
\end{equation}
with eigenvalues $\lambda_l $. Similarly, the eigenfunctions $\psi_l$ of $\tilde{H}$ are the eigenfunctions of the symmetric normalized Laplacian operator $\tilde{L} = I - \tilde{H}$, where $I$ is the identity operator. The normalized Laplacian $\tilde{L}$ is positive semidefinite and therefore has real non-negative eigenvalues $0 = \lambda_{1} < \lambda_{2} \leq \ldots \leq 2$.

The eigenfunctions $\phi_l$ correspond to temporal patterns of the dynamical flow and capture different timescales of the system induced by the probability measures in (\ref{eq:probMeasure}) (as will be illustrated in Sect.\ \ref{sect:exp}) acting as filters on $L^2(\nu)$. Moreover, since the function $J(\gamma)$ with $\gamma \in C^0(\Img;\real^m)$ from Theorem~\ref{thm:B} is in $L^2(\nu)$, there exist expansion coefficients $c_l \in \mathbb{R}^m$ such that we can expand $J(\gamma)$ in the $\{ \phi_l \}$ eigenfunction basis:
\begin{equation}\begin{gathered}\label{eq:reconstruction}
J(\gamma) = \sum_{l=1}^{\infty} c_l \psi_l = \sum_{l=1}^{\infty} c_l v^{\frac{1}{2}} \phi_l, \\
c_l := \langle \psi_l, J(\gamma) \rangle_{\nu} = \langle v^{\frac{1}{2}} \phi_l, J(\gamma) \rangle_{\nu} = \int_{\Stat} v^{\frac{1}{2}}(\pi) \phi_l(\pi)^* J(\gamma)(\pi) d\nu(\pi) \\
= \int_{X} v^{\frac{1}{2}} (p_x) \phi_l(p_x)^* J(\gamma)(p_x) d\mu(x)  = \int_{X} v^{\frac{1}{2}}(p_x) \phi_l(p_x)^* A_{\Delta t}(\gamma\circ f)(x) d\mu(x).
\end{gathered}
\end{equation}

Using (\ref{eq:reconstruction}), for $\gamma = \gamma^{(n)}$ with $\gamma^{(n)} : \real^{d} \to \real^{d}$ from (\ref{eq:defGammaMoments}) and $J(\gamma) = \mathbb{E}_n$, the moments of the probability distributions $p_x$ can be written as
\begin{equation}
\label{eq:rec	onstructionMoments}
\mathbb{E}_n(p_x) = \sum_{l=1}^{\infty} c_l v^{\frac{1}{2}}(p_x) \phi_l (p_x), \quad
c_l := \langle v^{\frac{1}{2}}\phi_l, \mathbb{E}_n \rangle_{\nu} = \int_{X} v^{\frac{1}{2}}(p_x) \phi_l(p_x)^* A_{\Delta t}(\gamma^{(n)} \circ f)(x) d\mu(x),
\end{equation}
where $c_l \in \mathbb{R}^d$ are the expansion coefficients used for the moment reconstruction. 

In Sect.\ \ref{sect:approx} we will consider the data-driven implementation of these quantities and of Theorems~\ref{thm:A} and \ref{thm:B}. We end this section by considering a special case when $\Stat$ has an additional geometric structure.

\subsection{Statistical manifolds}\label{subsect:statisticalManifolds}

A case of particular interest is when the collection $\Stat$ of probability measures is itself a manifold, thus justifying the name of \textit{statistical manifold}, i.e., a manifold where each point is a probability measure. Statistical manifolds and their properties are studied in the field of information geometry \citep{AmariNagaoka07, Nielsen18} using techniques from differential geometry. If the probability measures are all absolutely continuous with respect to a reference measure, say $\tau$, then $\Stat$ can be locally parameterized with a coordinate system $\boldsymbol{\theta}=\{\theta^1, \ldots, \theta^W\}$ inducing a parametric family of probability densities $\rho( \cdot ,\boldsymbol{\theta})$ on $ \mathbb{ R }^d $ with respect to $\tau$.  To track the statistical distances, i.e., divergences, between probability distributions locally near a point $p\in\Stat$, one needs to define a Riemannian metric $g_{p}$. There is a  canonical Riemannian metric, called the Fisher information metric, which measure the amount of information between two PDFs. It is an inner product on the tangent spaces of $\Stat$ defined by the natural basis of tangent vectors $\left \{\frac{\partial}{\partial \theta^1}, \ldots , \frac{\partial}{\partial \theta^W} \right \}$, given by
\begin{equation}\label{eq:fisherInfoMetric}
	g_{kl}(\boldsymbol{\theta}) = - \mathbb{E} \left[ \frac{\partial\ \mbox{log}\ \rho(\cdot,\boldsymbol{\theta})}{\partial \theta^k}\frac{\partial\ \mbox{log}\ \rho(\cdot,\boldsymbol{\theta})}{\partial \theta^l} \right].	
\end{equation} 
The Fisher information metric as defined above is positive definite and transforms as a type $ (0,2) $ tensor on $ \mathcal{ S } $ under changes of coordinate system $ \boldsymbol{\theta} $. Given any two tangent vectors $u = \sum_{k=1}^W u^k \frac{ \partial\; }{\partial \theta^k } $ and $v = \sum_{k=1}^W v^k \frac{ \partial\; }{\partial \theta^k } $  in $ T_p\Stat$, their inner product wrt $ g_p $ becomes $\langle u, v\rangle_p=\sum_{k,l=1}^W g_{kl}(\boldsymbol{\theta}) u^k v^l$.

The symmetric $W \times W$ matrix with elements $g_{kl}$ from~\eqref{eq:fisherInfoMetric}, $\mathcal{I}(\boldsymbol{\theta})=[g_{kl}]$, is called the Fisher information matrix (FIM) and is a positive definite matrix. The natural divergence on statistical manifolds associated with FIM is the Fisher information distance. For infinitesimal small changes in the probability distributions (i.e., $\boldsymbol{\theta}_{x'}=\boldsymbol{\theta}_{x}+d\boldsymbol{\theta}, $ for $p_x, p_{x'} \in \mathcal{S}$), the Fisher information distance can be expressed using the quadratic differential form, i.e.,
\begin{equation*}
	ds^2 = d \boldsymbol{\theta}^T \mathcal{I}(\boldsymbol{\theta}) d \boldsymbol{\theta} =  \sum_{k,l=1}^W g_{kl} d\theta^k \otimes d\theta^l.
	\label{eq:FisherDist}
\end{equation*}
Approximations of the parametric Fisher information distance can be achieved by well-known nonparametric divergences in the literature, such as the Kullback-Leibler (KL) divergence, the Hellinger distance or the cosine distance \citep{KassVos11}. From the family of $f$-divergences, the Hellinger distance satisfies all metric properties, and it has been shown to outperform the symmetric KL divergence (Jeffreys divergence) and Bhattacharyya distance (at least in the Gaussian case) due to the fact that the latter divergences do not obey the triangle inequality \citep{AbouEtAl12}. Its metric properties also allow for isometric embeddings in reproducible kernel Hilbert spaces \citep{AbouEtAl12}. These properties of the Hellinger distance provide an additional argument for our choice of divergence in the probability space, and the Hellinger distance remain a good choice of a metric as an approximation of the Fisher information distance even for the case when $\Stat$ has the additional manifold structure.

\section{Numerical approximations}\label{sect:approx}

Though we introduced our framework in the continuous case, in practice we work with discrete versions of the continuous-time dynamical systems described above. For the discrete case, we make the following assumption.

\begin{Assumption}\label{asmptn:data}
	There is a sequence of states $x_{i} = \Psi_{ i \,\delta t}( x_0 )$ for some initial state $x_0 \in X $ and sampling interval $ \delta t $, and a time-ordered sequence of measurements $ y_i = f( x_i ) $. The sampling interval $ \delta t $ is such that the discrete-time dynamical system $(X, \Psi_{\delta t}, \mu)$ is also ergodic.
\end{Assumption}

The assumption on ergodicity of $\mu$ will be key to ensuring that our results of data analysis converge in the limit of large data ($N \to \infty$ and $\delta t \to 0$). In most practical situations, the invariant measure $\mu$ is unknown to us. Instead, given a dataset consisting of $N$ samples, $\mu$ is approximated using the discrete sampling measure
\[ \mu_N = \frac{1}{N}\sum_{i=0}^{N-1} \delta_{x_i}, \]
the average of Dirac-$\delta$ measures supported on a trajectory $\{x_0,\ldots,x_{N-1}\}$. By the ergodicity of $\mu$ (Assumption \ref{asmptn:1}), for $\mu$-a.e.\ (almost everywhere) $x_0\in X$, the discrete sampling measures $\mu_N$ converge weakly to $\mu$. This means that for a set of initial points $x_0$ with $\mu$ measure equal to one,
\[ \int_X \kappa d\mu = \lim_{N\to\infty} \int_X \kappa d\mu_N = \lim_{N\to\infty} \frac{1}{N}\sum_{i=0}^{N-1} \kappa(x_i), \quad \forall \kappa \in C^0(X). \]
Ergodicity is a property of the system that is often implicitly assumed in data-driven studies of dynamical systems. It provides a justification for the principle that the global statistical properties of an observable $f : X \to \real^d$ wrt $\mu$ can be obtained from a time series for $f$, namely, $f(x_0), \ldots, f(x_{N-1})$.

The second discretization required is for the measures $p_x$. These are the push-forward measures under $g_x$ of the Lebesgue probability measure $\lambda$ on $T$ \eqref{eq:probMeasure}. The measure $\lambda$ will now be discretized by the measure $\lambda_R= \frac{1}{R}\sum_{r=0}^{R-1} \delta_{-r \delta t}$ on $T$, where $\Delta t = R\delta t$, and $R$ is the number of samples within an embedding window. The interval $T = [-\Delta t, 0]$ has been discretized as $\{ -r\delta t : 0\leq r < R \}$. As a result we have,
\begin{equation}
\label{eq:phat}
\hat{p}_x := g_{x*} \lambda_R = \frac{1}{R}\sum_{r=0}^{R-1} \delta_{ f \left( \Psi_{-r \delta t} (x) \right)}, 
\end{equation}
where $\hat{p}_x = \hat{p}(x)$ is the mapping of $x \in X$ under the map $\hat{p} : X \to \Prob_c(\Y)$. Just as with the measure $\nu$, $p$ and $\hat{p}$ also push the sampling measure $\mu_N$  into the discrete measures $\nu_N = p_* \mu_N$ and $\hat{\nu}_N = \hat{p}_* \mu_N$, respectively. $\nu_N$ and $\hat{\nu}_N$ are measures on the collections $\Stat$ and $\hat{\Stat}$ respectively, where $\hat{\Stat} := \hat{p}(X) = \{ \hat{p}_x \mid x \in X \}$. Note that both $\Stat$ and $\hat{\Stat}$ are contained in $\Prob_c(\Y)$. In the data-driven approximation scheme, the space $L^2(\mu)$ will be approximated by $L^2(\mu_N)$, and $L^2(\nu)$ by $L^2(\nu_N)$ and $L^2(\hat{\nu}_N)$. If the set $X$ is not a fixed point, then for $\mu$-a.e.\ $x_0$, these three spaces are isomorphic to $\cmplx^N$, and the functions in these spaces are $N$-dimensional vectors. The inner product on $L^2(\nu_N)$ is given by
\[ \langle \psi, \psi' \rangle_{\nu_{N}} = \frac{1}{N} \sum_{i=0}^{N-1} \psi(p_{x_i})^* \psi'(p_{x_i}), \quad \forall \psi, \psi' \in L^2(\nu_N), \]
where the last inner product is the usual inner product on $\cmplx^{N}$. $L^2(\mu_N)$ and $L^2(\hat{\nu}_N)$ carry analogous inner products.

Under the kernel density estimation, $\hat{p}_x = \hat{p}(x)$ is mapped into
\begin{equation}\label{eqn:def:rhohat}
 \hat{\rho}_{x}(y) := K(\hat{p}_x)(y) = \int_{\Y} k(y,y') d\hat{p}_x(y') = \frac{1}{R}\sum_{r=0}^{R-1} k\left( y, f \left( \Psi_{-r \delta t} (x) \right) \right), \quad \forall y\in \Y .
\end{equation}

Finally, we choose the reference measure $\DataMes$ on $\Y$ to be a discrete measure supported on $Q$ points $\{ z_q : q =0,\ldots, Q-1 \}$. The squared Hellinger distance becomes
\begin{equation}\label{eqn:def:Hellinger}
d_H^2\left( \hat{p}^{\DataMes}_x, \hat{p}^{\DataMes}_{x'}  \right) \stackrel{ \text{by \eqref{eq:Hellinger}}} {\longeq} \int_{\Y} \left[ \sqrt{K(\hat{p}_x)} - \sqrt{K(\hat{p}_{x'})}  \right]^2 d \DataMes  \stackrel{ \text{by \eqref{eqn:def:rhohat}} } {\longeq} \frac{1}{Q} \sum_{q=0}^{Q-1} \left[ \sqrt{\hat{\rho}_x (z_q)} - \sqrt{\hat{\rho}_{x'}(z_q)}  \right]^2. 
\end{equation}
We will overuse notation and use $d_H^2$ to also denote the distance induced on the set $\hat{\Stat} := \hat{p}(X)$ by the squared Hellinger distance in \eqref{eqn:def:Helling_data}, namely,
\begin{equation}\label{eqn:def:Helling_data}
d_H^2\left( \hat{p}_x, \hat{p}_{x'}  \right)  := d_H^2\left( \hat{p}^{\DataMes}_x, \hat{p}^{\DataMes}_{x'}  \right) = \frac{1}{Q} \sum_{q=0}^{Q-1} \left[ \sqrt{\hat{\rho}_x (z_q)} - \sqrt{\hat{\rho}_{x'}(z_q)}  \right]^2.
\end{equation}

\textit{Remark.} A typical choice for $\DataMes$ is any set of $Q$ points $\Y$ with an independent and identical distribution wrt some measure $\bar{\DataMes}$. As $Q\to\infty$, $\DataMes$ would converge weakly to $\bar{\DataMes}$. Another choice is $\DataMes = f_* \mu_Q$. This has the added advantage that it requires no extra points of evaluation other than on the original data points. This relates our work to kernel mean embedding techniques \cite{MuandetEtAl17,SriperumbudurEtAl2011} that have been used to embed probability measures into a Hilbert spaces of functions, called  Reproducing Kernel Hilbert Spaces (RKHS). This Hilbert structure provides additional tools, such as orthogonal projections, and allows to use pointwise evaluations as bounded linear functionals. We however do not use the RKHS aspect of the technique.

The infinite-dimensional space $L^2(\nu)$ will be approximated by the $N$-dimensional space $L^2(\hat{\nu}_N)$, the operators defined in Sect.~\ref{subsect:Hellinger} become $N\times N$ matrices. We will denote the matrix versions of all the operators using boldface notation. The kernel integral operator $G$ will be approximated by a $N\times N$ matrix $\DG{N} = \{ \DG{N}_{ij} \}$ acting on $L^2(\hat{\nu}_N)$, as
\begin{equation}\label{eqn:def:G_matrix}
\DG{N}_{i,j} = k_H \left( \hat{p}_{x_i}, \hat{p}_{x_j} \right) \stackrel{ \text{ by \eqref{eqn:def:kH}}} {\longeq} \exp \left( - \frac{1}{\epsilon} d_H^2\left( \hat{p}_{x_i}, \hat{p}_{x_j}  \right)  \right), \quad i,j = 0, \ldots, N-1.
\end{equation}
We have similar to the continuous case,
\begin{equation}\label{eqn:def:P_matrix}
\begin{gathered}
\Dq{N}  := \DG{N} \bm{1}, \quad \DtildeG{N}_{i,j} :=  \frac{ \DG{N}_{i,j} } {\Dq{N}_i \Dq{N}_j}, \quad \DtildeG{N}_{i,j} = \tilde{k}_H \left( \hat{p}_{x_i}, \hat{p}_{x_j} \right),\\
\Dv{N} :=  \DtildeG{N} \bm{1}, \quad \DaH{N}_{i,j} := \frac{\DtildeG{N}_{i,j}}{\Dv{N}_i} = \frac{1}{\Dv{N}_i } \frac{ \DG{N}_{i,j}}{ \Dq{N}_i \Dq{N}_j }, \quad \DaH{N} := \DV{N}^{-1} \DtildeG{N}.
\end{gathered}
\end{equation}
Let $\DV{N}$ be the diagonal matrix $\DV{N} := \diag\left( \Dv{N}_0, \ldots, \Dv{N}_{N-1} \right)$. Then note that $\DaH{N}$ is a Markov matrix of transition probabilities and is similar to the symmetric normalized matrix $\DtildeH{N}$ below:
\[ \DtildeH{N} := \DV{N}^{\frac{1}{2}} \DaH{N} \DV{N}^{-\frac{1}{2}} = \DV{N}^{\frac{1}{2}} \DV{N}^{-1} \DtildeG{N} \DV{N}^{-\frac{1}{2}} = \DV{N}^{-\frac{1}{2}} \DtildeG{N} \DV{N}^{-\frac{1}{2}} . \]
Since $\DtildeH{N}$ is a normalized symmetric matrix, it has a complete eigenbasis consisting of orthonormal vectors $\{ \bm{\psi}_{l} \}_{l=1}^{N}$, satisfying 
\begin{equation}
\DtildeH{N} \bm{\psi}_l = (1 - \bm{\lambda}_l) \bm{\psi}_l,
\end{equation}
with eigenvalues $1 - \bm{\lambda}_{l}$. The eigenvectors $\bm{\psi_l}$ are orthonormal with respect to the standard inner product $\langle \bm{\psi_k}, \bm{\psi_l} \rangle_{\hat{\nu}_N} = \frac{1}{N} \sum_{i=0}^{N-1} \bm{\psi_{k,i}} \bm{\psi_{l,i}} = \delta_{kl}$. Moreover, since $\DaH{N}$ and $\DtildeH{N}$ are related by a similarity transformation, $\DaH{N}$ has the same spectrum as $\DtildeH{N}$, and we have
\begin{equation}\label{eq:phiNs}
	\DaH{N} \bm{\phi}_{l} = (1 - \bm{\lambda}_{l}) \bm{\phi}_{l}, \quad \bm{\phi}_{l} := \Dv{N}^{-\frac{1}{2}} \bm{\psi}_{l}, \quad l=1, \ldots, N.
\end{equation}
The asymmetric random walk Laplacian matrix associated with the random walk matrix $\DaH{N}$ is $\bm{L} = \bm{I} - \DaH{N} = \bm{I} - \DV{N}^{-1} \DtildeG{N}$, and it has the same eigenvectors $\bm{\phi_l}$ as $\DaH{N}$:
\begin{equation}\label{eq:phiNsLaplacian}
	\bm{L} \bm{\phi}_{l} = \bm{\lambda}_{l} \bm{\phi}_{l}, \quad l=1, \ldots, N.
\end{equation}

$\bm{L}$ has the same eigenvalues $\bm{\lambda}_l$ as the symmetric normalized Laplacian matrix $\bm{\tilde{L}} = \bm{I} - \DtildeH{N}$ which is positive semidefinite and therefore has non-negative eigenvalues $0 = \bm{\lambda}_{1} < \bm{\lambda}_{2} \leq \ldots \leq \bm{\lambda}_{N} \leq 2$. $\bm{\tilde{L}}$ has the same eigenvectors as $\DtildeH{N}$, $\bm{\tilde{L}} \bm{\psi}_l = \bm{\lambda}_l \bm{\psi}_l$.  The Laplacian eigenvectors associated with the lowest eigenvalues $\bm{\lambda}_l$ vary slowly on the graph, i.e., two vertices that are connected by an edge with a large weight will have similar values of the leading eigenvectors \cite{ShumanEtAl13}.

The top eigenvector $ \bm{\psi}_1$ of $\DtildeH{N}$ coincides with $\Dv{N}^{\frac{1}{2}}$ and corresponds to the eigenvalue $1 - \bm{\lambda}_1 = 1$. This follows from the fact that since $\DaH{N}$ is a Markov operator, the eigenvector corresponding to eigenvalue $1$ is the constant vector $\bm{\phi}_1 = \bm{1}$. The eigenvectors $\bm{\phi}_l$ are orthonormal with respect to the $L^2(\omega_N)$ inner product, where the measure $\omega_N$ is absolutely continuous wrt the sampling measure $\hat{\nu}_N$ and has density $\bm{v} = \bm{\psi}_1^2 $:
\begin{equation*}
\langle \bm{\phi_k}, \bm{\phi_l} \rangle_{\omega_N} = \sum_{i=0}^{N-1} \bm{v}_i \bm{\phi_{k,i}} \bm{\phi_{l,i}} = \sum_{i=0}^{N-1} \bm{v}_i (\Dv{N}_i^{-\frac{1}{2}} \bm{\psi_{k,i}}) (\Dv{N}_i^{-\frac{1}{2}} \bm{\psi_{l,i}}) = \sum_{i=0}^{N-1} \bm{\psi_{k,i}} \bm{\psi_{l,i}} = \delta_{kl}.
\end{equation*}

We have outlined a summary of the entire numerical procedure to obtain the  $\bm{\lambda}_{l}$ and $\bm{\phi}_{l}$ in Algorithm 1. 

\begin{table}[ht!]
	\begin{center}
		\normalsize
		\begin{tabular}{p{16cm}}
			\hline
			\rule{0pt}{3ex}	\textbf{\large{Algorithm 1: Feature extraction (temporal patterns)}}\\
			\hline
			\rule{0pt}{3ex}	\textbf{Input}: $\Delta t$ -- embedding window\\ 
			\hspace{1.3cm} $R$ -- number of delay-coordinates, so that the sampling interval is $\delta t = \Delta t / R$ \\ 
			\hspace{1.3cm} $\{y_i\}_{i=-R}^{N-1}$ -- time series of observations, equal to $f\left(\Psi_{i\delta t} (x_0) \right)$ for an initial state $x_0\in X$ \\
			\hspace{1.3cm}  $\{ z_q \}_{q=0}^{Q-1} \in \Y$ -- collection of $Q$ evaluation points for density estimation\\
			\hspace{1.3cm} $\epsilon$ -- Gaussian kernel parameter\\ 
			\hspace{1.3cm} $M$ -- spectral resolution parameter \\
			\rule{0pt}{3ex}	\textbf{Output}: $\Phi=\{ \bm{\phi}_{l} \}$ -- eigenvectors (temporal patterns)\\
			\hline
			\rule{0pt}{3ex}	1: For each $i =0 \ldots, N-1$ and $q = 0, \ldots, Q-1$, evaluate $\hat{\rho}_{x_i}(z_q)$ using \eqref{eqn:def:rhohat}.\\
			\rule{0pt}{3ex}	2: Compute the pairwise Hellinger distances $d_H^2\left( \hat{p}_{x_i}, \hat{p}_{x_j}  \right) $ using \eqref{eqn:def:Helling_data}. \\
			\rule{0pt}{3ex}	3: Construct the $N\times N$ matrix $\DG{N}$ from \eqref{eqn:def:G_matrix} with kernel bandwidth parameter $\epsilon$.\\
			\rule{0pt}{3ex}	4: Compute the Markov matrix $\DaH{N}$ and the diagonal matrix $\DV{N}$ from \eqref{eqn:def:P_matrix}. \\
			\rule{0pt}{3ex}	5: Compute the leading $M$ eigenvectors $\bm{\psi}_{l}$ of $\DtildeH{N}$ corresponding to the largest eigenvalues $1 - \bm{\lambda}_{l}$. \\
			\rule{0pt}{3ex}	6: Compute the leading $M$ Laplacian eigenvectors $\bm{\phi}_{l}$ of $\bm{L}$ corresponding to the smallest eigenvalues $\bm{\lambda}_{l}$. \\
			\hline
		\end{tabular}
	\end{center}
	\label{tab:alg}
\end{table} 

\subsection{Moment reconstruction} 

Using Theorems~\ref{thm:A} and \ref{thm:B} and \eqref{eq:reconstruction}, we can expand the moments of the distributions in the $\{ \bm{\phi}_l \}$ eigenvector basis. The moments can be reconstructed (Algorithm 2) using only the leading $M$ eigenvectors and the expansion coefficients $\bm{c}_l \in \mathbb{R}^d$ as follows: 
\begin{equation}\label{eq:reconstrEigBasis}
\hat{\mathbb{E}}_n(p_{x_i}) := \sum_{l=1}^{M} \bm{c}_{l} \Dv{N}_{i}^{\frac{1}{2}} \bm{\phi}_{l,i}, 
\end{equation}
where $\Dv{N}_{i} = v(\hat{p}_{x_i})$, and
\begin{equation}\label{eq:expansionCoef}
\bm{c}_{l} := \int_X v^{\frac{1}{2}}(\hat{p}_x) \phi_l(\hat{p}_x)^* A_{\Delta t}\left( \gamma^{(n)}\circ f \right)(x) d\mu_N(x) = \frac{1}{N} \sum_{i=0}^{N-1} \Dv{N}_{i}^{\frac{1}{2}} \bm{\phi}_{l,i}^* \mathbb{E}_n(\hat{p}_{x_i}).
\end{equation}

\begin{table}[ht!]
	\begin{center}
		\normalsize
		\begin{tabular}{p{16cm}}
			\hline
			\rule{0pt}{3ex}	\textbf{\large{Algorithm 2: Moments reconstruction}}\\
			\hline
			\rule{0pt}{3ex}	\textbf{Input}: $\Phi=\{ \bm{\phi}_{l} \}$ -- eigenvectors\\ 
			\hspace{1.3cm} $\hat{p}_{x_i}$ -- the estimated probability distributions for the embedding windows\\ 
			\hspace{1.3cm} $\Dv{N}$ -- the density vector\\
			\rule{0pt}{3ex}	\textbf{Output}: $\hat{\mathbb{E}}_n(p_{x_i})$ -- reconstructed moments\\
			\hline
			\rule{0pt}{3ex}	1: Compute the true (first few) moments of each probability measure $\mathbb{E}_n(\hat{p}_{x_i})$.\\
			\rule{0pt}{3ex}	2: Compute the expansion coefficients $\bm{c}_l$ using (\ref{eq:expansionCoef}). \\
			\rule{0pt}{3ex}	3: Reconstruct the moments $\hat{\mathbb{E}}_n(p_{x_i})$ from (\ref{eq:reconstrEigBasis}) using only the leading $M$ eigenvectors $\bm{\phi}_{l}$. \\ 
			\hline
		\end{tabular}
	\end{center}
	\label{tab:alg:moment}
\end{table} 

\subsection{Continuous extensions} 
\label{subsect:continuousExt}

In the following, we will need to make more clear the dependence of the data-driven quantities used previously in this section on the parameters $N$ and $R$, and we will add these parameters as subscripts. For example, the vectors $\bm{v}$ and $\bm{\phi}_l$ from \eqref{eqn:def:P_matrix} and \eqref{eq:phiNs} will be denoted as $\bm{v}_{N,R}$ and $\bm{\phi}_{N,R,l}$, respectively. The dependence on $R$ comes from the fact that these quantities are constructed using the map $\hat{p}$ from \eqref{eq:phat} which depends on $R$.  

The eigenvectors $\bm{\phi}_{l}$ from \eqref{eq:phiNs}, denoted as $\bm{\phi}_{N,R,l}$, are $N$-dimensional vectors and are interpreted as functions on the set $\{ \hat{p}_{x_i} : i=0,\ldots,N-1 \}$. These functions can be continuously extended to the space $\Prob_c(Y)$, which contains both $\mathcal{S}$ and $\mathcal{\hat{S}}$. This is done in a manner similar to \eqref{eqn:def:P_matrix}. We use a boldface notation for vectors, and have dropped the boldface for their continuous extensions. In the equations below $\pi, \pi' \in \Prob_c(Y)$:
\begin{equation}\label{eq:outOfSampleExtension}
\begin{gathered}
q_{N,R}(\pi) := \frac{1}{N} \sum_{i=0}^{N-1} k_H(\pi, \hat{p}_{x_i}), \quad \tilde{k}_H(\pi, \pi') := \frac{k_H(\pi, \pi')}{ q_{N,R}(\pi) q_{N,R}(\pi') }, \quad v_{N,R}(\pi) := \frac{1}{N} \sum_{i=0}^{N-1} \tilde{k}_H(\pi, \hat{p}_{x_i} ), \\
h(\pi, \pi') = \frac{k_H(\pi, \pi')}{ v_{N,R}(\pi)}, \quad \tilde{h}(\pi, \pi') := \frac{\tilde{k}_H(\pi, \pi')}{ v_{N,R}(\pi)^{1/2} v_{N,R}(\pi')^{1/2} } = \frac{ k_H(\pi, \pi') }{ v_{N,R}(\pi)^{1/2} q_{N,R}(\pi) q_{N,R}(\pi') v_{N,R}(\pi')^{1/2} },\\
\phi_{N,R,l}(\pi) := \bm{\lambda}_{l}^{-1/2} \frac{1}{N} \sum_{i=0}^{N-1} \tilde{h}(\pi, \hat{p}_{x_i}) \bm{\phi}_{N,R,l,i}, \quad \text{where}\ \bm{\phi}_{N,R,l,i} = \bm{\phi}_{l,i} = \bm{\phi}_l(\hat{p}_{x_i}).
\end{gathered}
\end{equation}

As a result of these definitions, there exists a continuous function $\phi_{N,R,l}: \Prob_c(Y) \to \mathbb{R}$ such that the $i$-th coordinate of the vector $\bm{\phi}_{N,R,l}$ is given by 
\[ \bm{\phi}_{l,i} = \bm{\phi}_{N,R,l,i} = \phi_{N,R,l}(\hat{p}_{x_i}), \quad i=0, \ldots, N-1. \]

Just as the functions $\phi_{N,R,l}$ are continuous extensions of the vectors $\bm{\phi}_{N,R,l}$, the functions $v_{N,R}$ and $q_{N,R}$ from \eqref{eq:outOfSampleExtension} are continuous extensions of the vectors $\bm{v}_{N,R}$ and $\bm{q}_{N,R}$, respectively.

\bigskip 
We now have the following theorem establishing the convergence of our data-driven methods.

\begin{theorem}\label{thm:data}
Let Assumptions \ref{asmptn:1} and \ref{asmptn:data} hold, and $v_{N,R}$ and $\phi_{N,R,l}$ be the data-driven functions as above. Then for every  $\gamma\in C^0(\Img;\real^m)$ and $\mu$-a.e.\ $x\in X$, we have
\begin{equation*}\begin{gathered}
J(\gamma)(p_{x}) = \int_{\Y} \gamma d p_{x}  = \lim_{M\to\infty} \lim_{N\to\infty} \lim_{R\to\infty} \sum_{l=1}^{M} \bm{c}_{N,R,l} v_{N,R}^{\frac{1}{2}}(p_x) \phi_{N,R,l}(p_x), \\
\bm{c}_{N,R,l} := \frac{1}{N} \frac{1}{R} \sum_{i=0}^{N-1} \sum_{r=0}^{R-1} \bm{v}_{N,R,i}^{\frac{1}{2}} \bm{\phi}_{N,R,l,i} \gamma(y_{i-r}),\quad \textnormal{where}\ \bm{c}_l \in \mathbb{R}^m. 
\end{gathered}\end{equation*}
\end{theorem}

\paragraph{Remark} The moments $\mathbb{E}_n$ of the distributions $p_x$ can be expanded in the $\{ \bm{\phi}_l \}$ eigenvector basis. If we consider only the subspace $\mbox{span}\{ \bm{\phi}_1, \ldots , \bm{\phi}_M \}$ spanned by the first leading $M$ eigenvectors, then we can approximate the moments using only this reduced basis. Results on a real-world time series show that a small number of eigenvectors is already sufficient to accurately approximate the moments (see Sect.\ \ref{sect:exp}). Out-of-sample extension methods such as the kernel analog forecasting method \citep{ZhaoGiannakis16} can be used to forecast future values of the eigenfunctions, and therefore also to predict the moments of the distributions $p_x$.

\paragraph{The role of dimensionality} While density estimation in high dimensions is known to pose numerical problems because of the curse of dimensionality (i.e., the number of samples needs to grow exponentially with the number of dimensions \citep{Scott08}), nonparametric multivariate density estimation such as kernel estimators have proven to perform well in low dimensions ($d \leq 3$). If the sampling density $\delta t \rightarrow 0$, nonparametric density estimation techniques will perform well even in higher dimensions, i.e., for higher-dimensional observables. In the numerical experiments presented in this paper we mostly consider the observable $f$ to be one- or two-dimensional, $f:X \rightarrow \real$ or $f:X \rightarrow \real^2$, and we will use 1D or 2D kernel density estimation techniques.

\paragraph{Strength and limitations} We proposed this framework for ergodic deterministic dynamical systems, however our experiments and additional initial results indicate that in practice the approach is also applicable to non-stationary and stochastic systems (such as RMM (Sect.~\ref{subsect:RMM})) with promising results. The probability measures capture temporal information on the dynamics, thus going beyond standard analysis in ambient data spaces, however part of this dynamical information is lost within the embedding windows themselves. 

\paragraph{Comparison with previous works} Our framework is related to the works of \cite{TalmonCoifman13} and \cite{LianEtAl15}. In \cite{TalmonCoifman13}, the time-varying PDFs are approximated by histograms, and the pairwise distances between them are computed using the Mahalanobis distance. Computation of these distances requires the estimation of time-window local covariance matrices between histograms, which can prove to be costly, especially if the features (histograms) are high-dimensional, i.e., a high number of histogram bins. When taking the features to be equal to the logarithm of the histograms, the local covariance matrix becomes related to the Fisher information matrix, thus revealing an implicit connection to the field of information geometry. The Fisher information matrix and the associated statistical manifold are explicitly employed in \cite{LianEtAl15}, where a parametric model of the underlying PDFs is assumed. More precisely, the observations at each time step are assumed to be drawn from a multivariate Gaussian distribution with time-evolving parameters. The Kullback-Leibler divergence is used to compute the distances between the PDFs on the statistical manifold, and to define a Gaussian kernel used further for dimension reduction with Diffusion Maps. In \cite{TalmonCoifman13} and \cite{LianEtAl15}, the dynamical systems are stochastic, i.e.,  dynamical noise is present, and the observations are assumed to be drawn from a time-varying probability density function. Empirical PDFs over time windows are subsequently introduced for estimation purposes. We work with deterministic dynamical systems where the probability measures are induced by trajectories of the dynamical system over specified time windows. Thus, even though methodologically our approach has certain aspects in common with the methods of \cite{TalmonCoifman13} and \cite{LianEtAl15}, there are differences in perspective since we consider how a deterministic dynamical system acts on probability densities of observables over time windows, rather than using such windows for estimation purposes in a stochastic setting. Our framework does allow for observational noise which is different from stochastic/dynamical noise. Another difference between their approach and ours is the way we use the empirical histograms to compute pairwise distances between states of the dynamical system. We use nonparametric multivariate density estimation to compute joint densities between the vector components, i.e., the vector-valued observables $f$, and assign pairwise distances between dynamical states using the Hellinger distance between those densities. On the other hand, \cite{TalmonCoifman13} assigns pairwise distances through the Mahalanobis distances of concatenated histograms of each vector component,  while \cite{LianEtAl15} assumes a parametric model for the generating PDFs, i.e., a multivariate Gaussian model with time-evolving local covariance matrices. 

\section{Proofs of theorems}\label{sect:proofs}

\subsection{Proofs of Theorems \ref{thm:A} and \ref{thm:B}}

The following basic result from measure theory will be needed. It is commonly referred to as the ``change of variable'' formula for integrals.

\begin{lemma}\label{lem:change_var_integ}
	Let $(A,\xi)$ be a measure space, and $\Transient:A\to B$ a measurable one-to-one map. Let $\chi\in L^1(B,\Transient_*\xi)$ and $\Transient_*\xi$ be the push-forward of the measure $\xi$ under $\Transient$ (see the diagram below).
	\[\begin{tikzcd}
		(A, \xi) \arrow{r}{\Transient}
		& (B, \Transient_*\xi) \arrow{r}{\chi}
		&\real^{m}
	\end{tikzcd}\]
	Then $\int_B \chi d(\Transient_*\xi) = \int_A (\chi\circ \Transient) d\xi$. 
\end{lemma}

This lemma is a basic result from Analysis \citep[e.g., see][Thm 1.6.9]{Durrett2019} and we will skip the proof. We next prove Theorem \ref{thm:B}.

\paragraph{Proof of Theorem \ref{thm:B}} Note that $d\lambda=\frac{dt}{\Delta t}$ and the right hand side (RHS) of \eqref{eqn:change_var} can be rewritten as 
\[
	\mbox{RHS}=\frac{1}{\Delta t} \int_{-\Delta t}^0 \left(\gamma \circ f(\Psi_t( x))\right)dt = \frac{1}{\Delta t} \int_{-\Delta t}^0 \left(\gamma \circ g_x(t) \right) dt = 	\int_{-\Delta t}^0 \left(\gamma \circ g_x(t)\right)\frac{dt}{\Delta t} = \int_{-\Delta t}^0 \left(\gamma \circ g_x \right)d\lambda. 
\]

Now invoke Lemma \ref{lem:change_var_integ} with the substitutions $A=[0,\Delta t]$, $B=\real^{d}$, $\xi=\lambda$, $\chi=\gamma$, $\Transient=g_x$ to get,
\[\mbox{RHS}= \int_{-\Delta t}^0 \left(\gamma \circ g_x \right)d\lambda = \int_{\Y} \gamma d(g_{x*}\lambda) = \int_{\Y} \gamma dp_x = J(\gamma)(p_x) = \mbox{LHS}.\]
This completes the proof of the theorem. \qed

\paragraph{Proof of Theorem \ref{thm:A}} The $n$-th moment $\mathbb{E}_n$ of $p_x$ can be rewritten in terms of the linear transformation $J(\gamma)$ as
\[ \mathbb{E}_n(p_x) = \int_{\Y} \gamma^{(n)}(y) dp_x(y) = J(\gamma^{(n)})(p_x). \]

It follows from this equation and by substituting $\gamma = \gamma^{(n)}$ in \eqref{eqn:change_var} that,
$$\mathbb{E}_n(p_x) = J(\gamma^{(n)})(p_x) =\Avg_{\Delta t}(\gamma^{(n)} \circ f)(x).$$  
This completes the proof of the theorem. \qed

\subsection{Proof of Theorem \ref{thm:D}} \label{sect:proof:D}

Since $\Img$ is the image of $f$, for every $x\in X$, the image of $g_x$ is contained in $\Img$ and therefore $p_x$, which is the push-forward of $\lambda$ under $g_x$, will have a support contained in $\Img$. This proves the first part of the claim. To prove the continuity of $p$, we have to show that for every continuous map $\zeta : \Y\to \real$ , the map $x \mapsto \int_{\Y} \zeta\ d p_x$ is continuous. So let $\epsilon>0$ and $x\in X$ be fixed. $\zeta$ restricted to the compact set $\Img$ is uniformly continuous and therefore there is a $\delta>0$ such that for $z,z'\in \Img$, if $d(z,z')<\delta$, then $|\zeta(z)-\zeta(z')|<\epsilon$. Now since $X$ is a compact set, and $\Psi_t$, $f$ are continuous maps, for $x'$ sufficiently close to $x$, we will have 
\[ \| g_x - g_{x'}\|_{C^0(T)} = \sup_{t\in T} \left| f\left( \Psi_t( x ) \right) - f\left( \Psi_t(x' ) \right) \right| < \delta. \]
Thus for every such $x'$ close to $x$, by the choice of $\delta$,
\[ \left| \zeta \circ g_x(t) - \zeta \circ g_{x'}(t) \right| < \epsilon, \quad \forall t\in T. \]
Invoking Lemma~\ref{lem:change_var_integ}, we get for every fixed $\zeta \in C^0(\Y)$ and every $y$ close to $x$,
\[\begin{split}
 \left| \int_{\Y} \zeta\ d p_x - \int_{\Y} \zeta\ d p_y \right| &= \left| \int_{\Y} \zeta\ d (g_{x*}\lambda) - \int_{\Y} \zeta\ d (g_{y*}\lambda) \right| = \left| \int_{T} (\zeta \circ g_x) d\lambda - \int_{T} (\zeta \circ g_y) d\lambda \right| \\
 & = \left| \int_{T} \left[ \zeta \circ g_x -  \zeta \circ g_y \right] d\lambda \right| < \int_{T} \left| \zeta \circ g_x -  \zeta \circ g_y \right| d\lambda < \epsilon.
 \end{split}\]
This proves the continuity of $p$ as claimed. Note that if $f$ is injective, the support of each $p_x$, which is the curve $\{f\left( \psi_t(x) \right) : t\in T\}$ are distinct, and hence the $p_x$ are necessarily distinct. If $k$ is strictly positive definite, then $K$ is injective by a result of Fukumizu \textit{et.\ al.\ }\citep[See][Thm 4]{FukumizuEtAl09}. Let $\iota_r :C^r_c(\Y, \real^+_0) \to C^0_c(\Y, \real^+_0) $ denote the canonical inclusion map. To prove the continuity of $\iota_r \circ K\circ p$, we need the following result.

\begin{lemma}[Mercer's theorem \cite{Mercer1909}]\label{lem:5k9h}
Let $\beta$ be a Borel measure on a first-countable, topological space, with compact support $Z$. Let $k:Z\times Z\to\real$ be a continuous, symmetric, strictly positive definite kernel on $Z$. Then there exists an orthonormal eigenbasis $\{e_i : i\in\num\}$ of $L^2(Z,\beta)$ of continuous functions such that 
\[ k(x,y) = \sum_{i\in\num} e_i(x) e_i(y), \quad \forall x,y \in Z. \]
Moreover, the convergence is absolute and uniform on $Z$.
\end{lemma}

Now let $x\in X$ and $x_n$ be a sequence of points converging to $x$. We will show that for every $\epsilon>0$, 
\[ \limsup_{n\to\infty} \left\| K(p(x_n)) - K(p(x)) \right\|_{C^0(\Y)} < 2\epsilon. \]
Let $\delta_1>0$ be fixed. Then for $x'$ sufficiently close to $x$, the Hausdorff distance of the images of the functions $g_x$ and $g_{x'}$ is less than $\delta_1$. Let $Z'$ be the closed $\delta_1$-neighborhood of the image of $g_x$. Thus for $x'$ close to $x$, the support of $p(x')$ will lie in $Z'$. Further, by Assumption \ref{asmptn:kernel}, there is a $\delta_2>0$ such that if $d(y,y')>\delta_2$, then $k(y,y')<\epsilon$. Let $Z$ be the closed $\delta_2$ neighborhood of $Z'$. Note that for every probability measure $p$ with support in $Z'$ and every $y \in \Y\setminus Z$,
\[ |(Kp)(y)| = \left| \int_{\Y} k(y,y') dp(y') \right|  = \left| \int_{Z'} k(y,y') dp(y') \right| \leq \int_{Z'} \left| k(y,y') \right| dp(y') \leq \int_{Z'} \epsilon dp(i) = \epsilon. \]
Therefore, since the supports of the $p(x_n)$ eventually lie in $Z'$, we have
\[ \limsup_{n\to\infty} \sup_{y\in \Y\setminus Z} \left| K(p(x_n)) - K(p(x))(y) \right| < 2\epsilon . \]
Thus it only remains to prove an inequality analogous to the one above, but for $y\in Z$. By Mercer's theorem, there are continuous functions $\{ e_i : i\in \num \}$ and $\lambda_i>0$ such that $k(y,y') = \sum_{i\in\num} e_i(y) e_i(y')$ uniformly and absolutely over $Z$. Thus there exists an $L\in\num$ such that 
\[  \left| k(y,y') - \sum_{i=1}^{L} e_i(y) e_i(y') \right| < \epsilon, \quad \forall y,y'\in Z . \]
Therefore for any probability measure $q$ with support in $Z'$, we have
\[ \sup_{y\in Z} \left| (Kq)(y) - \sum_{i=1}^{L} e_i(y) \int e_i(y') dq(y') \right| < \epsilon . \]
Now replacing $q$ by $p(x)$ and $p(x_n)$ respectively, gives
\[ \sup_{y\in Z} \left| K(p(x_n)) - K(p(x))(y) \right|  < \sum_{i=1}^{L} e_i(y) \left[ \int e_i(y') dp_x(y') - \int e_i(y') dp_{x_n}(y')  \right] + 2\epsilon . \]
Thus, applying $\limsup_{n\to\infty} $ on both sides give
\[\begin{split}
\limsup_{n\to\infty} \sup_{y\in Z} \left| K(p(x_n)) - K(p(x))(y) \right| &< \limsup_{n\to\infty} \sum_{i=1}^{L} \sup_{y\in Z} e_i(y) \left[ \int e_i(y') dp_x(y') - \int e_i(y') dp_{x_n}(y')  \right] + 2\epsilon \\
&= \sum_{i=1}^{L} \sup_{y\in Z} e_i(y) \limsup_{n\to\infty} \left[ \int e_i(y') dp_x(y') - \int e_i(y') dp_{x_n}(y')  \right] + 2\epsilon = 2\epsilon.
\end{split}\]
where we have used the fact that $p(x_n)$ converges weakly to $p(x)$. This completes the proof of continuity of $\iota_r \circ K\circ p$. Its injectivity follows from the injectivity of all these three maps. 

We will now show that $\iota_{\DataMes} : C^0_c(\Y, \real^+_0) \to \Prob^+(\Y;\DataMes)$ is a continuous map wrt the Hellinger distance. So let $\rho, \rho' \in C^0_c(\Y, \real^+_0)$. Since $\rho$ has compact support, it has a bounded range, $[a,b]$. If $\left\| \rho - \rho' \right\|_{C^0(\Y)} < \epsilon$, then by the continuity of the square-root function on the interval $[a,b]$, $\left\| \sqrt{\rho} - \sqrt{ \rho' } \right\|_{C^0(\Y)} < \epsilon$. Thus $d_H( \iota_{\DataMes} (\rho), \iota_{\DataMes}(\rho') ) < \epsilon)$. Since $\epsilon$ was arbitrary, this proves the continuity of $\iota_{\DataMes}$ at each point  $\rho \in C^0_c(\Y, \real^+_0)$. The proof of injectivity of $\iota_{\DataMes}$ is trivial and is left to the reader. This completes the proof of Theorem \ref{thm:D}. \qed

\subsection{Proof of Theorem \ref{thm:E}} \label{sect:proof:E}

To prove that a commutative diagram is true, it is necessary and sufficient to prove that the commuting relations in the smallest loops hold. We will now break down the commuting diagram in Theorem \ref{thm:E} into its smallest commutative components. First consider the relations
\[
\begin{tikzcd} X \arrow{r}{f} \arrow{dr}[swap]{f'} & Y \arrow{d}{D} \\ \  & Y' \end{tikzcd} \quad
\begin{tikzcd} X \arrow{r}{g} \arrow{dr}[swap]{g'} & C^0(T;Y) \arrow{d}{D} \\ \  & C^0(T;Y') \end{tikzcd} \quad
\begin{tikzcd} X \arrow{r}{g} \arrow{dr}[swap]{p} & C^0(T;Y) \arrow{d}{\Lambda} \\ \  & \Prob_c(Y) \end{tikzcd} \quad
\begin{tikzcd} X \arrow{r}{g'} \arrow{dr}[swap]{p'} & C^0(T;Y') \arrow{d}{\Lambda'} \\ \  & \Prob_c(Y') \end{tikzcd} \quad
\]
The first two relations follow from the definitions of $f'$ and $g'$. The next two follow from the definition in \eqref{eq:probMeasure}. Next consider the relations
\[
\begin{tikzcd} C^0(T;Y) \arrow{r}{\Lambda} \arrow{d} {D\circ} & \Prob_c(Y) \arrow{d}{D_*} \\ C^0(T;Y') \arrow{r}{\Lambda'} & \Prob_c(Y') \end{tikzcd} \quad
\begin{tikzcd} \Prob_c(Y) \arrow{d}{D_*} \arrow{r}{K} & C^r(Y;\real_0^+) \\ \Prob_c(Y') \arrow{r}{K'} & C^r(Y';\real_0^+) \arrow{u}{\circ D} \end{tikzcd} \quad
\begin{tikzcd} C^r(Y;\real_0^+) \arrow{r}{\iota_{\alpha}} & \Prob(Y, \alpha) \arrow{d}{D_*}[swap]{\cong} \\ C^r(Y';\real_0^+) \arrow{u}{\circ D} \arrow{r}{\iota_{D_*\alpha}} & \Prob(Y', D_*\alpha) \end{tikzcd} 
\]
The first relation is already explained in Sect.~\ref{subsect:Hellinger}. To prove the second relation, take some $\pi\in \Prob_c(Y)$. Then, for every $y\in Y$
\[\begin{split}
((K'D_* \pi)\circ D)(y) &= (K'D_* \pi)(Dy) = \int_{Y'} k'(Dy, z) d(D_* \pi)(z) \\
&= \int_{Y} k'(Dy, Dy') d\pi(y'),  \quad \mbox{by Lemma~\ref{lem:change_var_integ}},\\
& = \int_{Y} k(y,y') d\pi(y') = (K \pi)(y).
\end{split}\]
Therefore, $(K'D_* \pi)\circ D$ = $K \pi$. Finally, to check the last relation, take a $\varrho \in C^r(Y';\real_0^+)$ and let $\beta:=\iota_{\alpha}( \varrho \circ D)$. To show that the two measures $D_*\beta$ and $\iota_{D_*\alpha}(\varrho)$ are equal, it is enough to show that their integrals with every continuous function with compact support $\chi \in C_c^0(Y')$ are equal. This holds because
%
\[ \int_{Y'} \chi d \left(D_*\beta \right) = \int_{Y} (\chi \circ D) d\beta = \int_{Y} (\chi \circ D) (\varrho \circ D) d\alpha = \int_{Y'} \chi \varrho d(D_*\alpha) = \int_{Y'} \chi d \left(\iota_{D_*\alpha}(\varrho)\right) . \]
This completes the proof of Theorem \ref{thm:E}. \qed

\subsection{Proof of Theorem \ref{thm:data}} \label{sect:proof:data}

As discussed in Subsect.~\ref{subsect:continuousExt}, we will use $R$ and $N$ in the subscripts to denote the dependence on the parameters $R$ and $N$. Similarly to $v_{N,R}$ and $q_{N,R}$, we denote the data-driven matrices $\DG{N}$ from \eqref{eqn:def:G_matrix} and $\DaH{N}$ from \eqref{eq:phiNs} as $\bm{G}^{(N, R)}$ and $\bm{H}^{(N, R)}$, respectively, since they are constructed using the map $\hat{p}$ which depends on $R$ as in \eqref{eq:phat}. If they are constructed in an analogous manner using the map $p$, the resulting matrices will be denoted as $\bm{G}^{(N)}$ and $\bm{H}^{(N)}$. For the former pair of matrices, the Hilbert space in question is $L^2(\hat{\nu}_N)$, while in the latter case it is $L^2(\nu_N)$. One similarly gets vectors $\bm{v}_{N,l}$ and $\bm{\phi}_{N,l}$ and their continuous extensions $v_{N,l}$ and $\phi_{N,l}$, instead of $\bm{v}_{N, R,l}$, $\bm{\phi}_{N, R,l}$, $v_{N, R,l}$ and $\phi_{N, R,l}$. The expansion coefficients $\bm{c}_{N, R,l}$ will be denoted as $\bm{c}_{N,l}$ if we use the map $p$ instead of $\hat{p}$.

Given a function/observable $\gamma\in C^0(\Img; \real^m)$, since $J(\gamma) \in \mathcal{L} \left( \Mes(M); \real^m \right)$, $J(\gamma)$ is an $m$-dimensional vector of continuous functions on $\Stat$. For simplicity and without loss of generality, in the proof we make the assumption that $m=1$. Using the $\phi_l$ basis from \eqref{eq:phis}, $J(\gamma)$ can be represented as an $L^2(\nu)$ function:
\[  J(\gamma)= \lim_{M\to\infty} \sum_{l=1}^{M} c_l v^{\frac{1}{2}} \phi_l, \quad c_l := \langle v^{\frac{1}{2}} \phi_l,  J(\gamma) \rangle_{\nu}. \]

Here, the inner product is taken between $\psi_l = v^{\frac{1}{2}}\phi_l$ and each component of $J(\gamma)$ wrt the measure $\nu$. Based on results from \cite{DasGiannakis_delay_Koop, DasGiannakis_RKHS_2018}, $J(\gamma)$ can be approximated from data in the $L^2(\nu)$ norm. The following limit holds in the $L^2(\nu)$ sense for $\mu$-a.e.\ $x$:
\begin{equation}\label{eqn:jvh6x} 
J(\gamma) = \lim_{M\to\infty} \lim_{N\to\infty} \sum_{l=1}^{M} \bm{c}_{N,l} v_{N}^{\frac{1}{2}} \phi_{N,l}, \quad \bm{c}_{N,l} := \langle \Dv{N}_{N}^{\frac{1}{2}} \bm{\phi}_{N,l},  J(\gamma) \rangle_{\nu_N} = \frac{1}{N} \sum_{i=0}^{N-1} \Dv{N}_{N,i}^{\frac{1}{2}} \bm{\phi}_{N,l,i} \left[ \int_{\Y} \gamma d p_{x_i} \right].
\end{equation} 

As mentioned above, the functions $v_{N}$ and $\phi_{N,l}$ are continuous extensions of the vectors $\bm{v}_{N}$ and $\bm{\phi}_{N,l}$ when using the map $p$ instead of $\hat{p}$. From the theory of graph Laplacians \citep[e.g.,][]{VonLuxburgEtAl08}, we have
\begin{equation}\label{eqn:ljkbnc9s} 
\lim_{N\to\infty} \sup_{\pi\in \Prob_c(Y)} \left| v_N(\pi) - v(\pi) \right| = 0, \quad \lim_{N\to\infty} \sup_{\pi\in \Prob_c(Y)} \left| \phi_{N,l}(\pi) - \phi_l(\pi) \right| = 0, \quad \forall \ l \in \num_0.
\end{equation} 

Since the measures $\lambda_R$ converge weakly to $\lambda$, their push-forwards under $g_x$ also converge for every $x\in X$, namely,
\begin{equation} \label{eqn:lnv093} 
\hat{p}_x := g_{x*} \lambda_R = \frac{1}{R} \sum_{r=0}^{R-1} \delta_{f\left( \Psi_{-r\delta t} (x) \right)} \xrightarrow{\ w\ } g_{x*} \lambda = p_x \quad \mbox{as } R\to \infty,
\end{equation}
where $\xrightarrow{\ w\ }$ denotes weak convergence of measures. Thus, for every $1\leq l\leq N$,
\begin{equation}
\label{eq:c_RNL}
\begin{split}
\bm{c}_{N,l} &= \frac{1}{N} \sum_{i=0}^{N-1} \Dv{N}_{N,i}^{\frac{1}{2}} \bm{\phi}_{N,l,i} \left[ \int_{\Y} \gamma d p_{x_i} \right] = \frac{1}{N} \sum_{i=0}^{N-1} \Dv{N}_{N,i}^{\frac{1}{2}} \bm{\phi}_{N,l,i} \left[ \lim_{R\to\infty }\int_{\Y} \gamma d \hat{p}_{x_i} \right] \\
& = \frac{1}{N} \sum_{i=0}^{N-1} \Dv{N}_{N,i}^{\frac{1}{2}} \bm{\phi}_{N,l,i} \left[ \lim_{R\to\infty } \frac{1}{R} \sum_{r=0}^{R-1} \gamma(y_{i-r}) \right] = \lim_{R\to\infty } \bm{c}'_{N, R,l}, \quad \bm{c}'_{N, R,l} := \frac{1}{N} \frac{1}{R} \sum_{i=0}^{N-1} \sum_{r=0}^{R-1} \Dv{N}_{N,i}^{\frac{1}{2}} \bm{\phi}_{N,l,i} \gamma(y_{i-r}).
\end{split}
\end{equation}

Next, since $X$ is compact, the image $\Img = f(X)$ is compact too, and thus by \eqref{eqn:lnv093},
\[ \lim_{R\to \infty} \sup_{y\in\Img} \left| (K p_x)(y) - (K\hat{p}_x)(y) \right| = 0, \]
for every $x\in X$. As a result, for every $1\leq i,j \leq N$,
\[ \bm{G}^{(N, R)}_{i,j} = \exp\left( -\frac{1}{\epsilon} \int_{Y} \left[ \sqrt{K \hat{p}_{x_i}} - \sqrt{K \hat{p}_{x_j} } \right]^2 d\alpha \right) \xrightarrow{R\to\infty} \exp\left( -\frac{1}{\epsilon} \int_{Y} \left[ \sqrt{K p_{x_i}} - \sqrt{K p_{x_j} } \right]^2 d\alpha \right) = \bm{G}^{(N)}_{i,j}. \]

Thus, the $N\times N$ matrix $\bm{G}^{(N, R)}$ converges to $\bm{G}^{(N)}$ as $R\to\infty$. It can be similarly shown that $\bm{H}^{(N, R)}$ converges to $\bm{H}^{(N)}$. Therefore, by the theory of perturbation of compact operators \citep[][Sect.~7]{BabuskaOsborn1991}, their eigenvectors and eigenvalues also converge. In particular, the functions and vectors resulting from the continuous extensions in \eqref{eq:outOfSampleExtension} converge uniformly, too. In summary, we have
\begin{equation}
\label{eqn:kjn03}
\lim_{R\to\infty} \sup_{\pi\in \Prob_c(Y)} \left| v_{N, R}(\pi) - v_N(\pi) \right| = 0, \quad \lim_{R\to\infty} \sup_{\pi\in \Prob_c(Y)} \left| \phi_{N, R,l}(\pi) - \phi_{N,l}(\pi) \right| =0, \quad \forall \ l \in \num_0.
\end{equation}

The first consequence of the uniform convergence in \eqref{eqn:kjn03} is that
\begin{equation} \label{eqn:jn94} 
\bm{c}_{N,l} = \lim_{R\to\infty} \bm{c}'_{N,R,l} = \lim_{R\to\infty} \bm{c}_{N,R,l}, \quad \bm{c}_{N,R,l} := \frac{1}{N} \frac{1}{R} \sum_{i=0}^{N-1} \sum_{r=0}^{R-1} \bm{v}_{N, R, i}^{\frac{1}{2}} \bm{\phi}_{ N, R, l,i} \gamma(y_{i-r}) , \quad \forall \ 1 \leq l \leq N.
\end{equation}

The parameter $R$ can now be inserted in \eqref{eqn:jvh6x} using \eqref{eqn:ljkbnc9s}, \eqref{eqn:lnv093} and \eqref{eqn:jn94} to get
\[\begin{split}
J(\gamma) &= \lim_{M\to\infty} \lim_{N\to\infty} \sum_{l=1}^{M} \bm{c}_{N,l} v_{N}^{\frac{1}{2}} \phi_{N,l} = \lim_{M\to\infty} \lim_{N\to\infty} \sum_{l=1}^{M} \left[ \lim_{R\to\infty} \bm{c}_{N,R,l} \right] \left[ \lim_{R\to\infty} v_{N,R}^{\frac{1}{2}} \right] \left[ \lim_{R\to\infty} \phi_{N,R,l} \right] \\
&= \lim_{M\to\infty} \lim_{N\to\infty} \lim_{R\to\infty } \sum_{l=1}^{M} \bm{c}_{N,R,l} v_{N, R}^{\frac{1}{2}} \phi_{N, R,l}.
\end{split}\]

This completes the proof of Theorem \ref{thm:data}. \qed

\section{Experiments}\label{sect:exp}

In this section, we illustrate our feature extraction and moment reconstruction framework on applications to three low-dimensional dynamical systems on: 1) an integrable ergodic flow on the 2-torus with two different dynamical regimes, Model I (Fig.\ \ref{fig:data}(a)) and Model II (Fig.\ \ref{fig:data}(b)), 2) the Oxtoby system on the 2-torus \citep{Oxtoby53} with a fixed point (Fig.\ \ref{fig:data}(c)), and 3) the Lorenz 63 system (Fig.\ \ref{fig:data}(d)). Each dynamical system is observed only through partial observations in the sense that the map $ f $ is not one-to-one. Some of these systems, and their associated eigenvectors, have been previously analyzed using different kernels and operators -- Models I and II on the 2-torus using the cone kernel \citep{Giannakis15}, and the Oxtoby system with a fixed point, Models I and II using the Koopman operator \citep{Giannakis19}. These methods relied on a full observation map. 

\begin{figure}
	\subfigure[2-torus (Model I)]{\includegraphics[trim = 0cm 0cm 0cm 0cm, clip, width=0.24\textwidth]{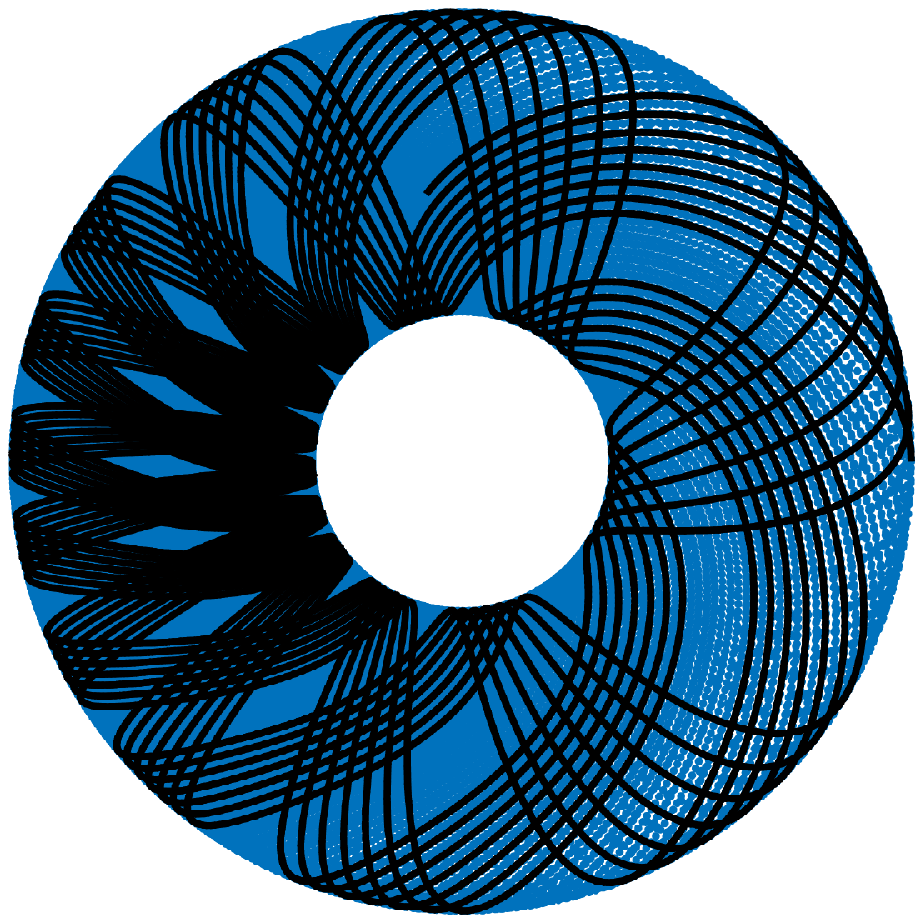}}
	\subfigure[2-torus (Model II)]{\includegraphics[trim = 2cm 2cm 2cm 2cm, clip, width=0.24\textwidth]{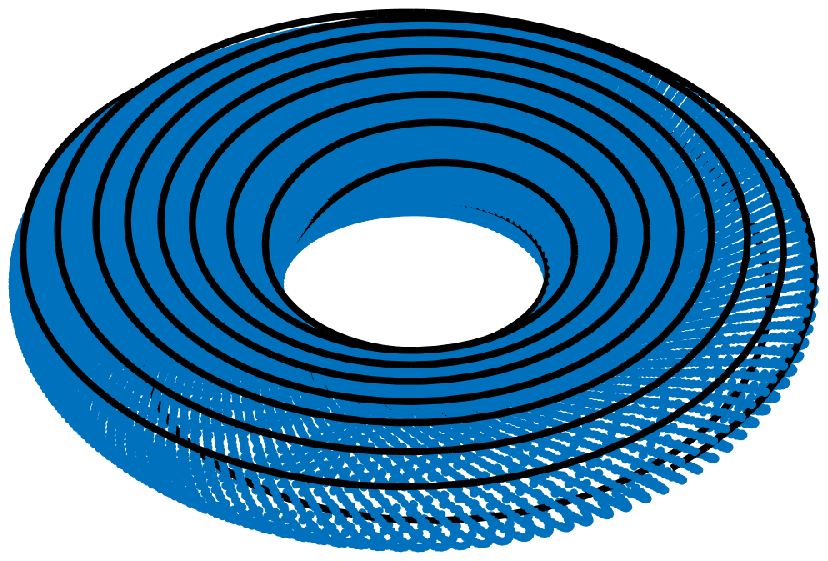}}
	\subfigure[2-torus with a fixed point]{\includegraphics[trim = 0cm 0cm 0cm 0cm, clip, width=0.24\textwidth]{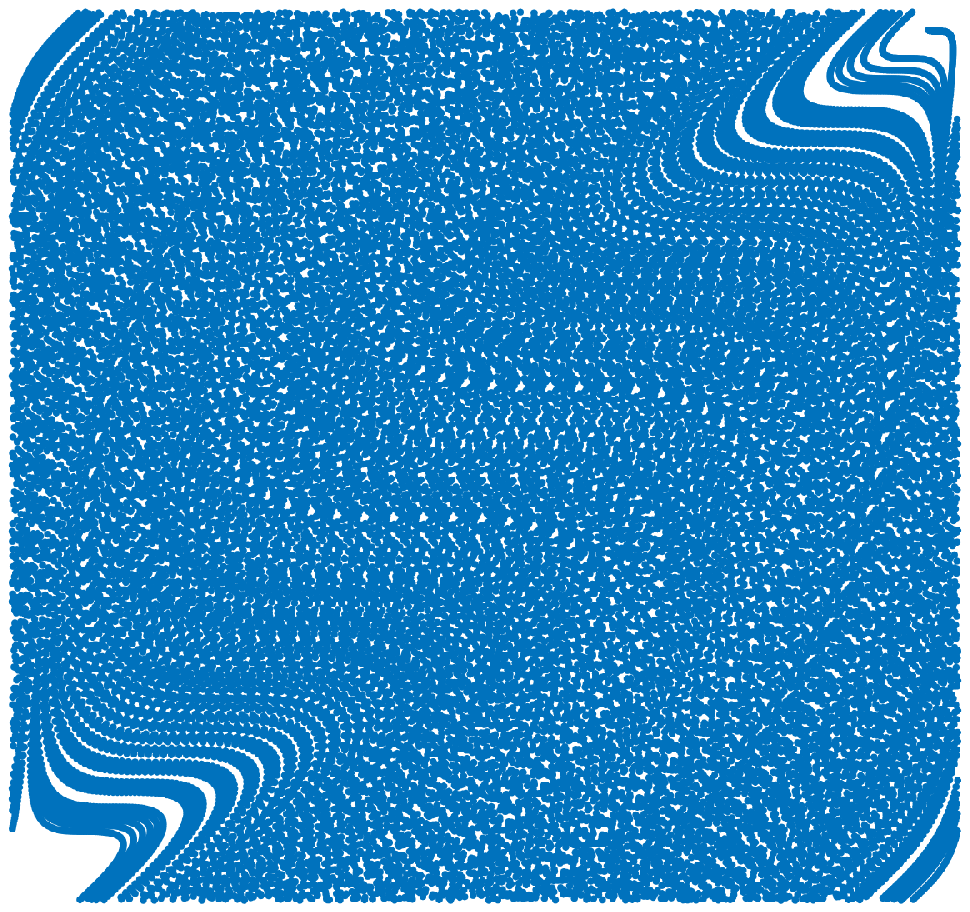}}
	\subfigure[Lorenz attractor]{\includegraphics[trim = 2cm 2cm 2cm 2cm, clip, width=0.24\textwidth]{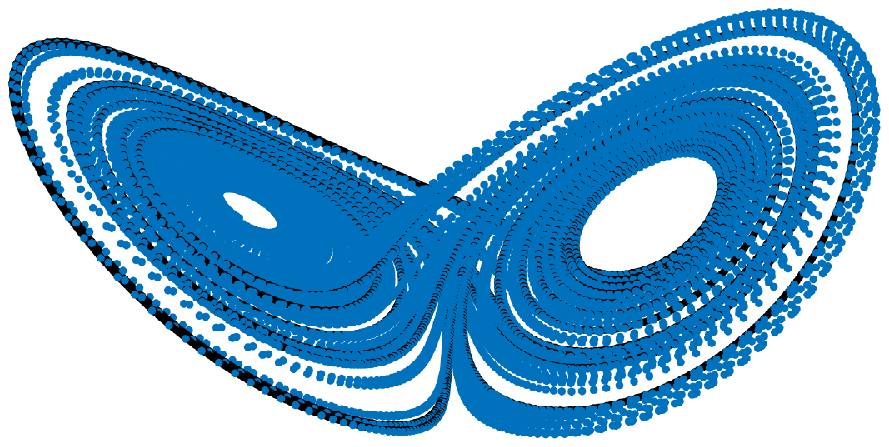}}
	\caption{The four dynamical systems used as examples.
	\label{fig:data}}
\end{figure}

\subsection{Integrable flow on the 2-torus}
\label{sect:2torus}

In our first example, we consider an ergodic dynamical system whose state space $ X $ is the 2-torus. Denoting the azimuthal and polar angles by $\vartheta^1$ and $\vartheta^2$, respectively, the vector field of this system is given by $ v = \sum_{i=1}^2 v^i \frac{ \partial\;}{ \partial \vartheta^i } $, where
\begin{equation}\label{eqn:2torus}
	v^1 = 1 + (1-\beta)^{1/2}\cos \vartheta^1, \hspace{0.5cm}v^2=\zeta(1-(1-\beta)^{1/2}\sin \vartheta^2),
\end{equation}
where $\zeta$ is an irrational angular frequency parameter. The speed variations of the flow are controlled through the parameter $\beta \in (0,1]$, such that the flow along the manifold speeds up or slows down when $\beta < 1$ (here $\beta=0.5$). We consider two dynamical models on the 2-torus with two different angular frequency, $\zeta=30^{1/2}$ and $\zeta=30^{-1/2}$, denoted Models I and II, respectively. 

The datasets for Models I and II were generated using $N=64,000$ samples at a sampling interval $\delta t=2\pi/S$, where $S=500$ controls the number of samples in each quasi-period (equivalent to approximately 128 periods). The full observation map is the standard embedding of the 2-torus in $\real^3$, that is, 

\begin{equation}
	\begin{gathered}
	\label{eqn:torusObs}
		F:X \mapsto \real^3, \quad F=(f^1,f^2,f^3), \\
		f^1(x)=(1+r^1\cos\vartheta^2(x))\cos\vartheta^1(x), \quad f^2(x)=(1+r^1\cos\vartheta^2(x))\sin\vartheta^1(x), \quad f^3(x)=r^2\sin \vartheta^2(x),
	\end{gathered}
\end{equation}
where both the azimuthal radius $r^1$ and polar radius $r^2$ are equal to $0.5$. 

All kernel density estimators use $Q=50$ evaluation points per dimension. In both models, we varied the parameter values and the results seem robust within certain ranges. See Appendix \ref{app:distanceDistrib} for more details.

\begin{figure}[ht]
	\includegraphics[trim = 1cm 0cm 0cm 0.5cm, clip, width=\textwidth]{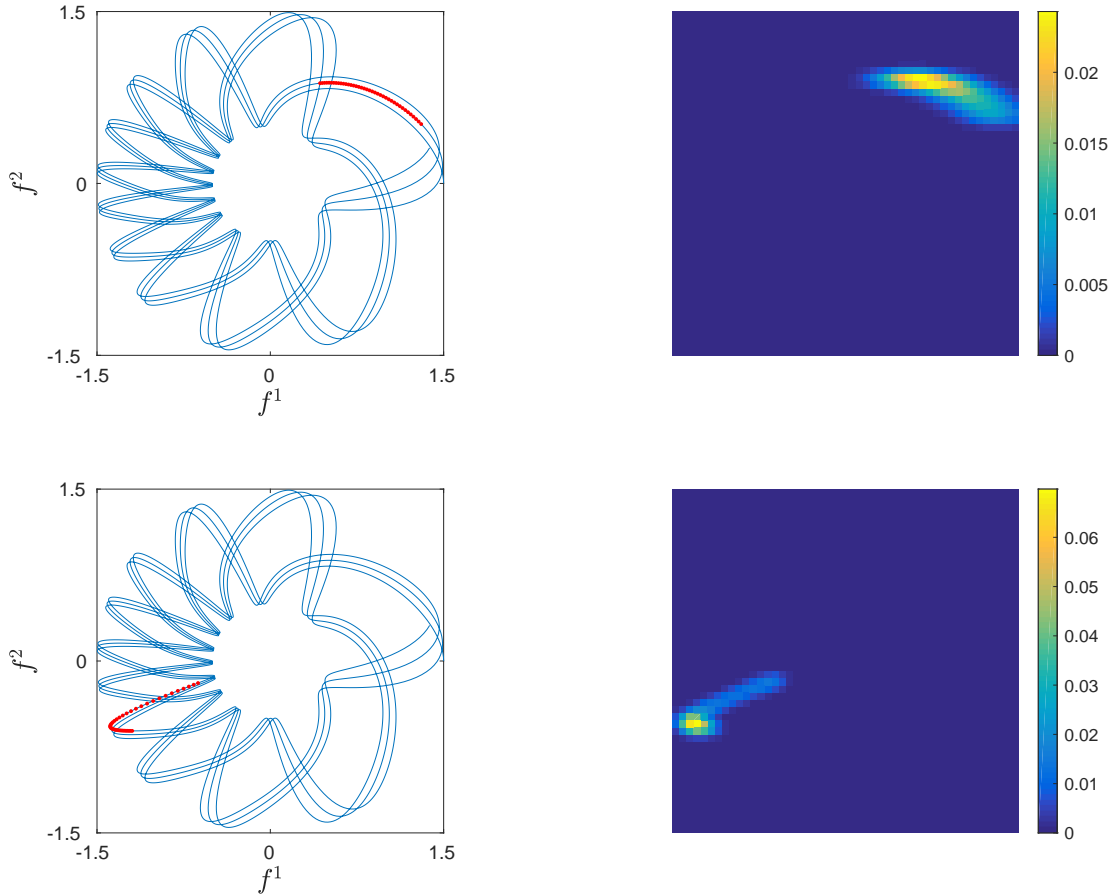}
	\caption{Examples of PDFs over trajectories of multivariate (2D) partial observables $f=(f^1,f^2)$ for the 2-torus \eqref{eqn:2torus}, Model I. The trajectories over an embedding window $R=40$ timesteps are shown in red on the left, and the corresponding PDFs estimated using kernel density estimation in 2D are shown on the right. All KDE estimators used $Q=50$ evaluation points per dimension.
	\label{fig:plotPDF}}
\end{figure}

\subsubsection{Model I}

The angular frequency of Model I was set to $\zeta=30^{1/2}$. Here, the partial observation map $f$ is given by $f=(f^1,f^2)$. Figure \ref{fig:plotPDF} shows two examples of PDFs over trajectories of multivariate (2D) observables for the 2-torus \eqref{eqn:2torus}, Model I. The eigenvectors are computed using (\ref{eq:phiNs}), and some of the most representative ones are shown in Fig.\ \ref{fig:2torusModelI_phis}. The embedding window in \eqref{eqn:def:rhohat} is set to $R=40$ timesteps, i.e., $\Delta t = R \delta t = 40 * 2 \pi / 500$. The parameters of the Diffusion Maps algorithm were set to $k=500$ nearest neighbors, the width of the Gaussian kerne $\epsilon=1$, and the normalization parameter $\alpha=1$. The normalization parameter $\alpha$ in the Diffusion Maps algorithm will be henceforth set to $\alpha = 1$, to fully decouple the geometry of the data from the density of the data \citep{CoifmanLafon06}. 

\begin{figure}[ht]
	\includegraphics[trim = 3.5cm 1.25cm 1cm 0.5cm, clip, width=\textwidth]{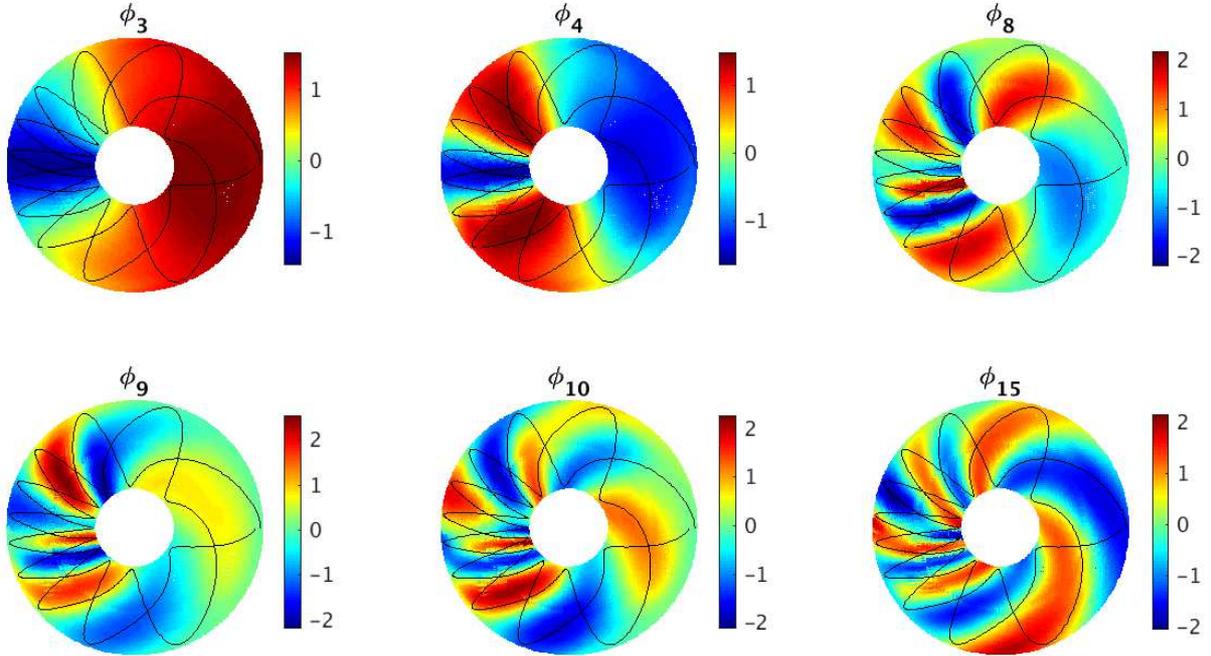}
	\caption{Representative eigenvectors of the 2-torus \eqref{eqn:2torus} with angular frequency $\zeta=30^{1/2}$. We used information only from the partially observed system $f=(f^1,f^2)$ from \eqref{eqn:torusObs}. The first eigenfunction is the constant vector of ones. A portion of the dynamical system trajectory is plotted in black for reference. 
	\label{fig:2torusModelI_phis}}
\end{figure}

The eigenvectors $\bm{\phi}_l$ in Fig.\ \ref{fig:2torusModelI_phis} capture different timescales of the dynamical flow. The probability measures introduced in (\ref{eq:probMeasure}) uncover temporal patterns of the underlying dynamical patterns despite having access to only partial observations of the system through $f$ instead of $F$ from \eqref{eqn:torusObs}. This is due to the fact that information on the dynamical evolution of the system is implicitly captured in the individual trajectories. The regions where the system evolves slowly (negative values along the horizontal axis in Fig.\ \ref{fig:2torusModelI_phis} corresponding to negative values of the azimuthal angle $\vartheta^1$) correspond to regions where the eigenvectors vary to a high degree. 

\subsubsection{Model II}

The angular frequency of Model II was set to $\zeta=30^{-1/2}$. Some of the most representative eigenvectors are shown in Fig.\ \ref{fig:2torusModelII_phis}. The embedding window in \eqref{eqn:def:rhohat} was set to $R=80$ timesteps. The parameters of the Diffusion Maps algorithm were set to $k=7,000$ nearest neighbors, the width of the Gaussian kernel $\epsilon=0.18$, and the normalization parameter $\alpha=1$.

\begin{figure}[ht]
	\includegraphics[trim = 3.5cm 1.5cm 1cm 0.5cm, clip, width=\textwidth]{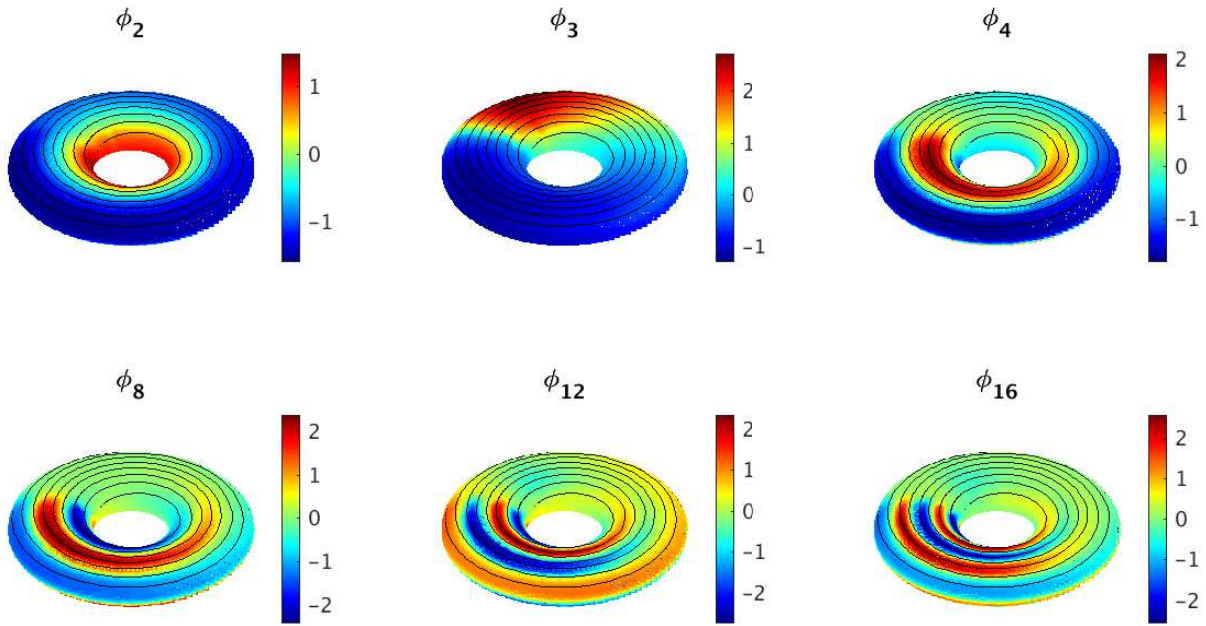}
	\caption{Representative eigenvectors of the 2-torus \eqref{eqn:2torus} with angular frequency $\zeta=30^{-1/2}$. We used information only from the partially observed system $f=(f^1,f^2)$ from \eqref{eqn:torusObs}. The first eigenfunction is the constant vector of ones. A portion of the dynamical system trajectory is plotted in black for reference. The torus is viewed in the 3D representation $F=(f^1,f^2, f^3)$ at $30^{\circ}$ azimuthal (horizontal) angle and $70^{\circ}$ vertical elevation.
	\label{fig:2torusModelII_phis}}
\end{figure}

The eigenvectors $\bm{\phi}_l$ capture different timescales of the dynamical system. The slow timescale on the 2-torus has time period $T=2 \pi / \min \{1, \zeta \}$, i.e., $T=2 \pi$ for Model I, and $T=2 \pi \times 30^{1/2}$ for Model II. This is equivalent to saying that the slow timescales happen along $\vartheta^1$ for Model I, and along $\vartheta^2$ for Model II. From Figs.\ \ref{fig:2torusModelI_phis} and \ref{fig:2torusModelII_phis} we see that the eigenvectors vary in directions transverse to the dynamical flow, and are able to capture the characteristic dynamical patterns of each model, with swirl patterns for Model I and azimuthal patterns for Model II. In the $\real^3$ standard embedding representation of the 2-torus, the slow timescales for Model II happen along the third dimension $f^3$. Our approach captures these slow timescales despite the fact that the system is observed only partially through $f=(f^1, f^2)$ without any information on $f^3$.

\subsubsection{Kernels based on Euclidean distances in the ambient data space}

For comparison, we also computed the eigenvectors directly in the full ambient data space using $k=10,000$ nearest neighbors, a standard Gaussian kernel with width $\epsilon=0.25$, and Diffusion Maps normalization with $\alpha=1$. We used the Euclidean distance to compute the pairwise distances. The eigenvectors for Models I and II are shown in Figs.\ \ref{fig:dataSpace1} and \ref{fig:dataSpace2}, respectively. The distances are computed using the fully observed system in $\real^3$, $F=(f^1, f^2, f^3)$. However, despite the fact that the models generating the data are different, the embeddings are identical for the same set of parameter values. Thus, the ambient data space is not capable of capturing through the eigenvectors the difference in the dynamical evolution of the two systems.

\begin{figure}
	\includegraphics[trim = 3.5cm 1.25cm 1cm 0.5cm, clip, width=\textwidth]{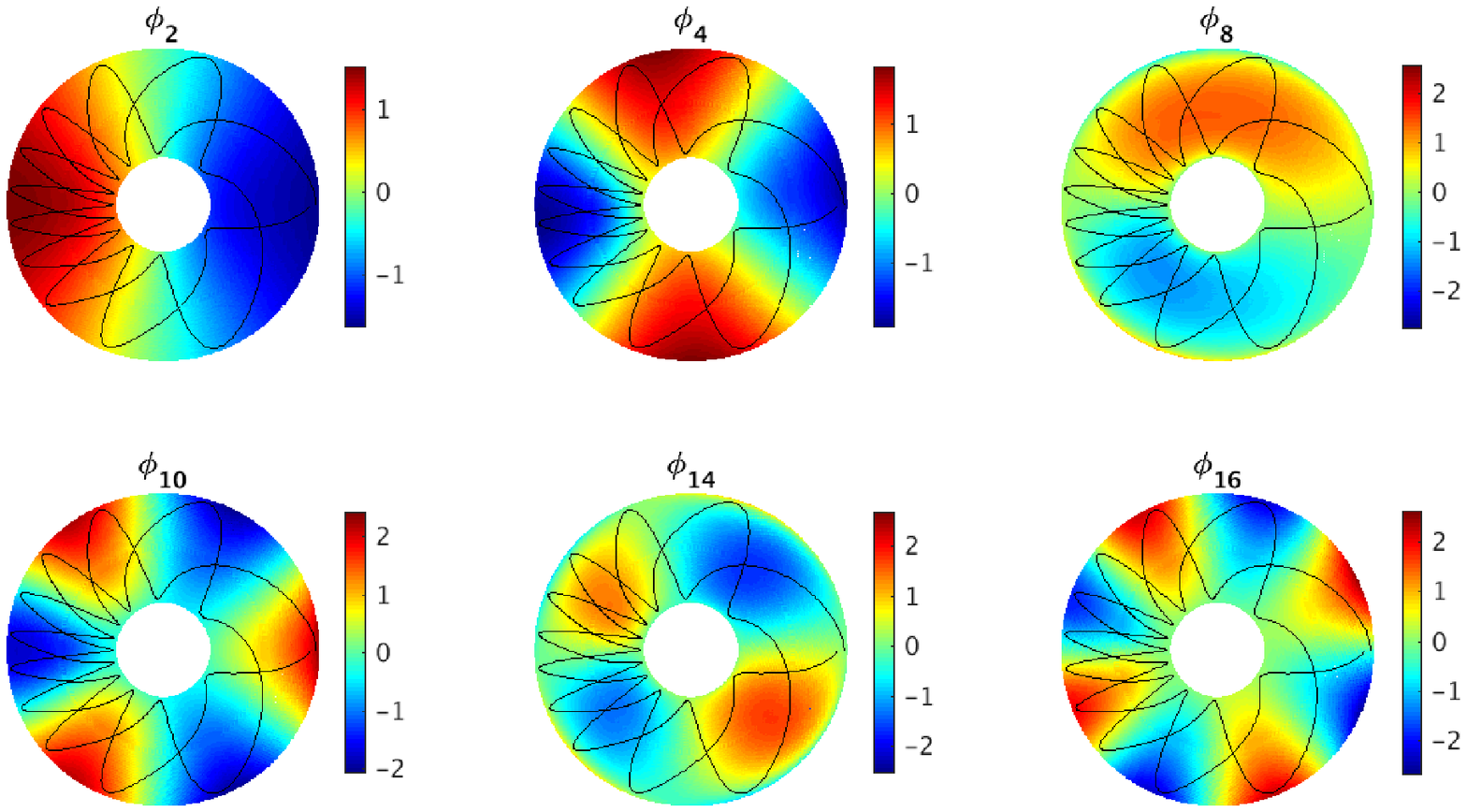}
	\caption{Representative eigenvectors of the 2-torus with angular frequency $\zeta=30^{1/2}$ in the ambient data space. Distances were computed using the full standard embedding representation $F=(f^1,f^2,f^3)$ \eqref{eqn:torusObs} of the 2-torus in $\real^3$.
	\label{fig:dataSpace1}}
\end{figure}

\begin{figure}
	\includegraphics[trim = 3.5cm 1.5cm 1cm 0.5cm, clip, width=\textwidth]{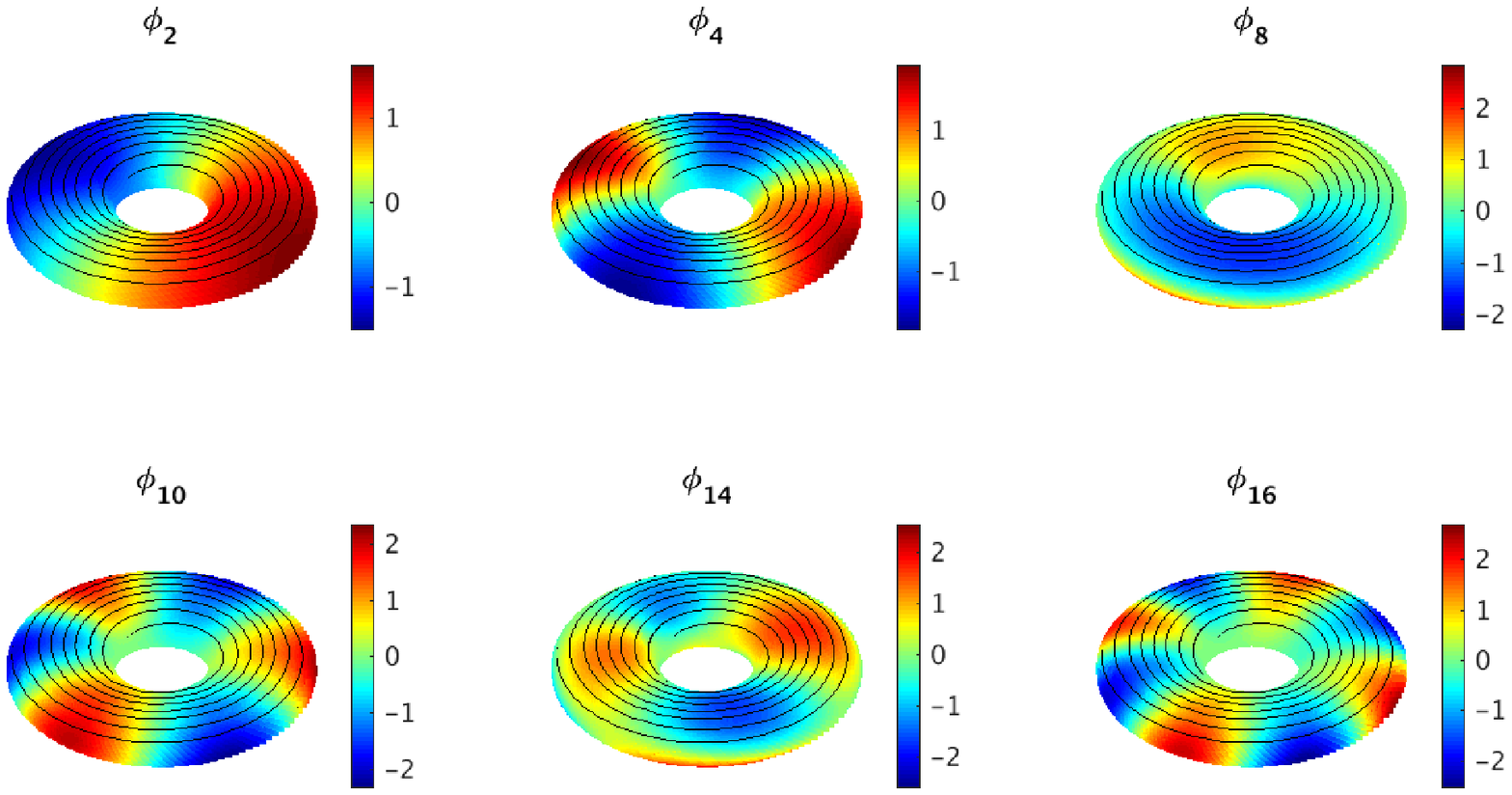}
	\caption{Representative eigenvectors of the 2-torus with angular frequency $\zeta=30^{-1/2}$ in the ambient data space. Distances were computed using the full standard embedding representation $F=(f^1,f^2,f^3)$ \eqref{eqn:torusObs} of the 2-torus in $\real^3$.
	\label{fig:dataSpace2}}
\end{figure}

\subsection{\label{sect:Oxtoby}Dynamics on the 2-torus with a fixed point}

For this second experiment, we consider the dynamical system on the 2-torus proposed by \cite{Oxtoby53} with a fixed point and the vector field $v=(v^1,v^2)$ given by
\begin{equation}
	v^1 = v^2 + (1-\zeta)(1-\cos \vartheta^2), \hspace{0.5cm}v^2=\zeta(1-\cos(\vartheta^1-\vartheta^2))
	\label{eq:2torusFixedPoint}
\end{equation}
where $\zeta$ is an irrational frequency parameter. The flow in (\ref{eq:2torusFixedPoint}) has a fixed point at coordinates $\vartheta^1=\vartheta^2=0$. Trajectories along this dynamical system pass by the fixed point at arbitrarily small distances, but they circumvent the fixed point by developing ``bumps'' around the fixed point. The standard (flat) embedding of the 2-torus in $\real^4$ is given by

\begin{equation}
	\begin{gathered}
	\label{eqn:obs2}
		F:X \mapsto \real^4, \quad  F=(f^1,f^2,f^3,f^4), \\
		f^1(x)=\cos \vartheta^1(x), \quad f^2(x)=\sin \vartheta^1(x), \quad f^3(x)=\cos \vartheta^2(x), \quad f^4(x)=\sin \vartheta^2(x).
	\end{gathered}
\end{equation}

\begin{figure}[ht]
	\includegraphics[trim = 2.5cm 0cm 1.5cm 0cm, clip, width=\textwidth]{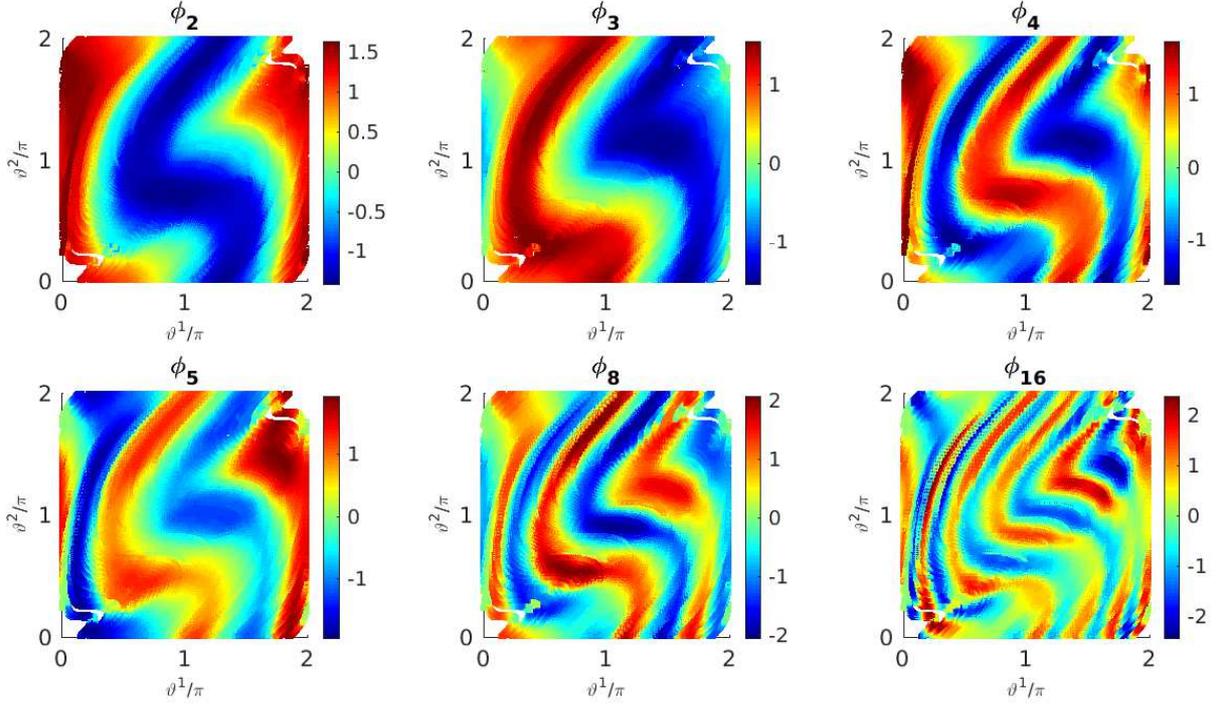}
	\caption{Leading eigenvectors of the 2-torus with a fixed point for the partially observed system $f=(f^1,f^2)$ from \eqref{eqn:obs2}. The first eigenfunction is the constant vector of ones.
	\label{fig:2torusFixedPoint_phis}}
\end{figure}

We generate $N=64,000$ points from this dynamical system for the frequency $\zeta=20^{1/2}$ and a timestep $\delta t=0.01$. The probability measures were estimated using KDE with an embedding window of $R=40$ timesteps and $Q=50$ evaluation points per dimension. The parameters of the Diffusion Maps algorithm were set to $k=3,000$ nearest neighbors, the width of the Gaussian kernel $\epsilon=1$, and the normalization parameter $\alpha=1$.

The leading eigenvectors for the system partially observed through $f=(f^1,f^2)$ instead of the full observational map $F$ from \eqref{eqn:obs2} are shown in Fig.\ \ref{fig:2torusFixedPoint_phis}. Their associated scatterplots, i.e., spatial patterns, follow the orbits of the dynamical flow, despite partial observation of the system. We performed the same analysis for all partially-observed systems through all 1D and 2D combination of observables of the full coordinate space $F=(f^1, f^2, f^3, f^4)$. The results are consistent for the different cases, and the eigenvectors capture mainly the slowly-varying timescales of the system, i.e., trajectories along the orbits of the flow. 

\subsection{Lorenz attractor}
\label{sect:lorenz}

In the third experiment, we consider the Lorenz 63 mathematical model initially proposed as a simple model for atmospheric convection, which consists of three ordinary differential equations:

\begin{equation}
	\label{eqn:l63}
	\frac{d\omega^1}{dt}=\sigma(\omega^2-\omega^1), \quad \frac{d\omega^2}{dt}=\omega^1(\rho-\omega^3)-\omega^2, \quad \frac{d\omega^3}{dt}=\omega^1 \omega^2-\beta \omega^3.
\end{equation}

Here $\omega^1,\omega^2,\omega^3$ are the system states, and $\sigma, \rho, \beta$ are the system parameters. We consider here the typical parameter values for the Lorenz system: $\rho=28, \sigma=10, \beta=8/3$. The embedding in $\real^3$ is given by

\begin{equation}
	\begin{gathered}
	\label{eqn:obs_l63}
	F:X \mapsto \real^3, \quad  F=(f^1,f^2,f^3), \\
	f^1(x)=\omega^1(x), \quad f^2(x)=\omega^2(x), \quad f^3(x)=\omega^3(x).
	\end{gathered}
\end{equation}

We generated $N=66,828$ points starting at the initial point $(0,1,1.05)$ for the time interval $\Delta T=[0,500]$, after having removed the first 150 transient points. The probability measures were estimated using KDE with an embedding window of $R=30$ timesteps and $Q=50$ evaluation points per dimension. The parameters of the Diffusion Maps algorithm were set to $k=2,000$ nearest neighbors, the width of the Gaussian kernel $\epsilon=0.32$, and the normalization parameter $\alpha=1$. The Lorenz 63 system is highly nonlinear and non-periodic, and feature extraction is therefore a challenging problem. 

The leading eigenvectors for the system partially observed through $f=(f^1,f^3)$ and $f=(f^1,f^2)$ are shown in Figs.\ \ref{fig:lorenz_x1x3_phis} and \ref{fig:lorenz_x1x2_phis}, respectively. The eigenvectors $\bm{\phi}_l$ capture different patterns of the slowly-varying timescales of the system, while faster-varying timescales emerge as we go deeper in the eigenfunction spectrum. We observe that some of the eigenvectors in Figs.\ \ref{fig:lorenz_x1x3_phis} and \ref{fig:lorenz_x1x2_phis} overlap (with sometimes only a change of sign), indicating that the framework proposed here recovers the underlying dynamics under different partial observables of the system. 

Figure \ref{fig:lorenz_x1x2_phis1D_2D} shows examples of one-dimensional time series and two-dimensional representations of the eigenvectors for the system partially observed through $f=(f^1, f^2)$. In this case, eigenfunction $\bm{\phi}_2$ represents the two wings of the Lorenz attractor, i.e., positive and negative values correspond to the left and right wing of the attractor, respectively (see also Fig.\ \ref{fig:lorenz_x1x2_phis}); eigenfunction $\bm{\phi}_3$ represents the variation within each wing with positive values corresponding to points further apart from the intersection of the wings, while negative values correspond to points closer to the intersection; and eigenfunction $\bm{\phi}_4$ represents a switch between the wings.

\begin{figure}[ht]
	\includegraphics[trim = 3cm 0cm 0.5cm 0cm, clip, width=\textwidth]{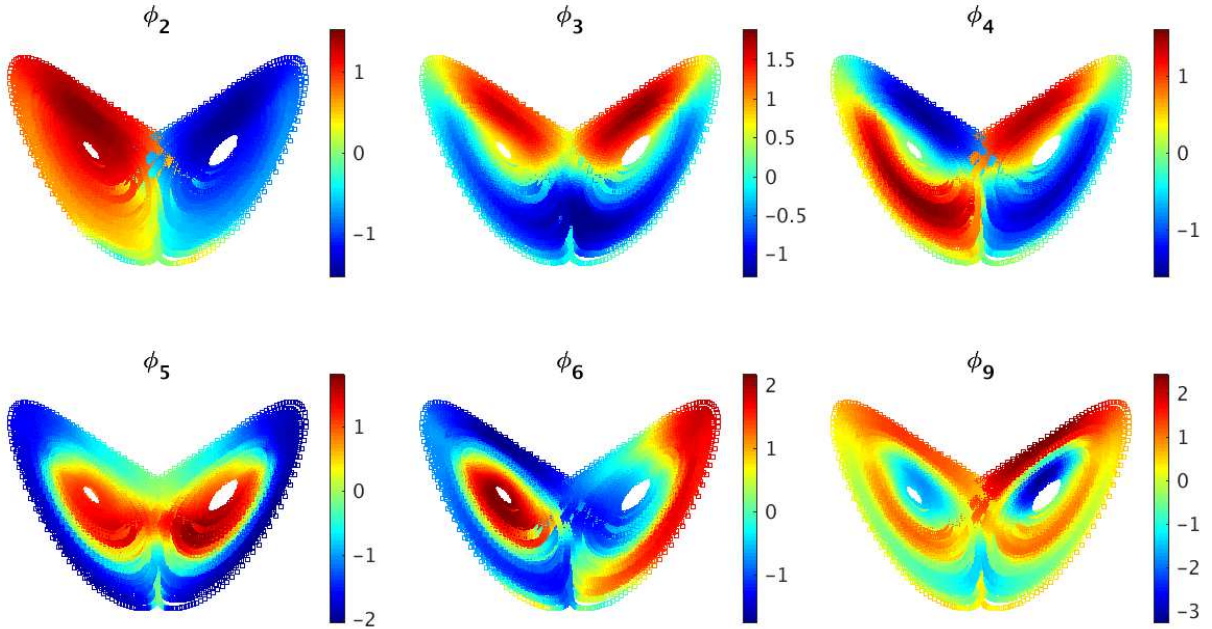}
	\caption{Leading eigenvectors of the Lorenz 63 system \eqref{eqn:l63} partially observed through $f=(f^1,f^3)$ from \eqref{eqn:obs_l63}. The first eigenfunction is the constant vector of ones.
	\label{fig:lorenz_x1x3_phis}}
\end{figure}

\begin{figure}[ht]
	\includegraphics[trim = 3cm 0cm 0.5cm 0cm, clip, width=\textwidth]{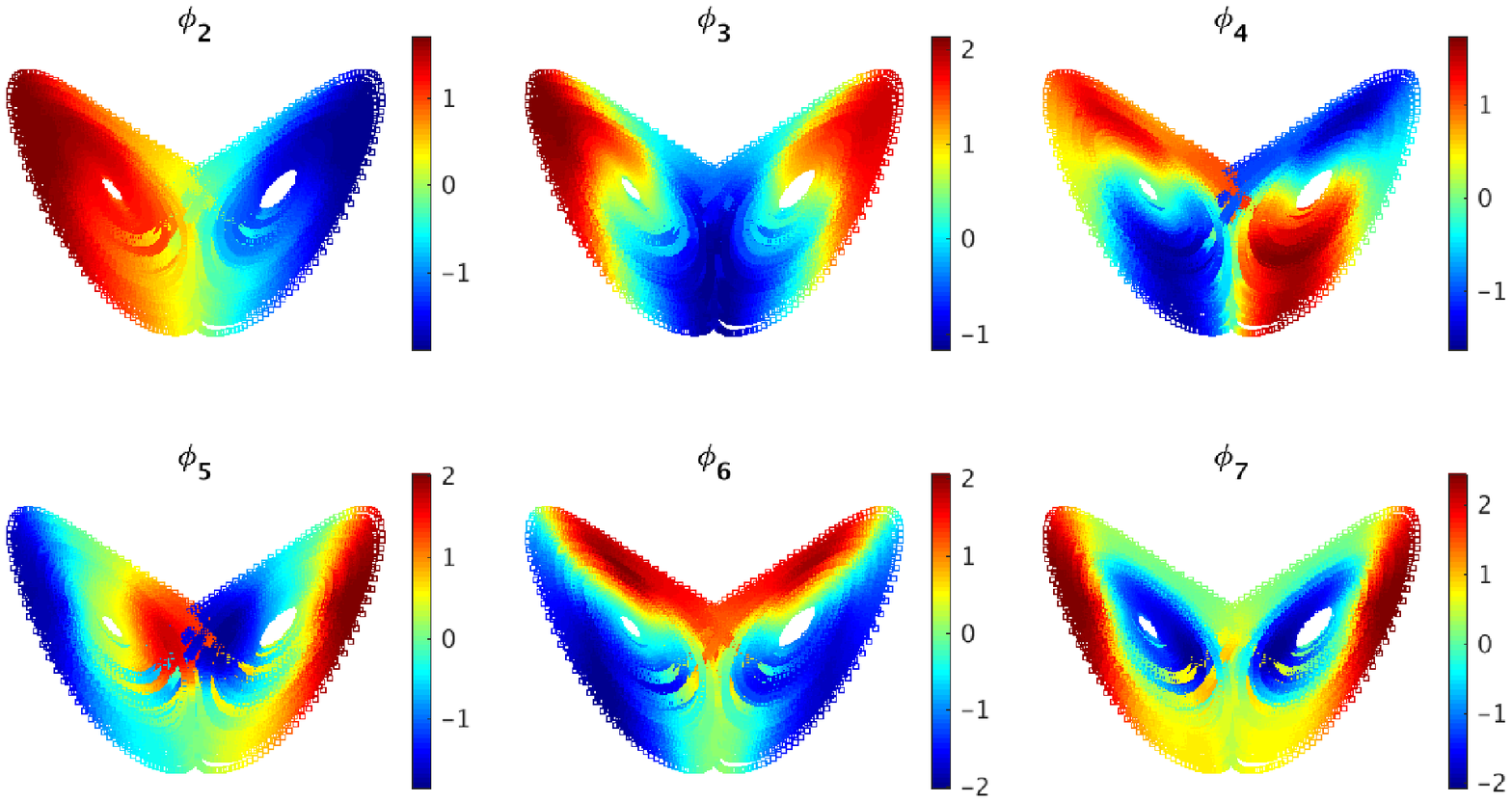}
	\caption{Leading eigenvectors of the Lorenz 63 system \eqref{eqn:l63} partially observed through $f=(f^1,f^2)$ from \eqref{eqn:obs_l63}. The first eigenfunction is the constant vector of ones. 
	\label{fig:lorenz_x1x2_phis}}
\end{figure}

\begin{figure}[ht]
	\includegraphics[trim = 2cm 0cm 2cm 0cm, clip, width=\textwidth]{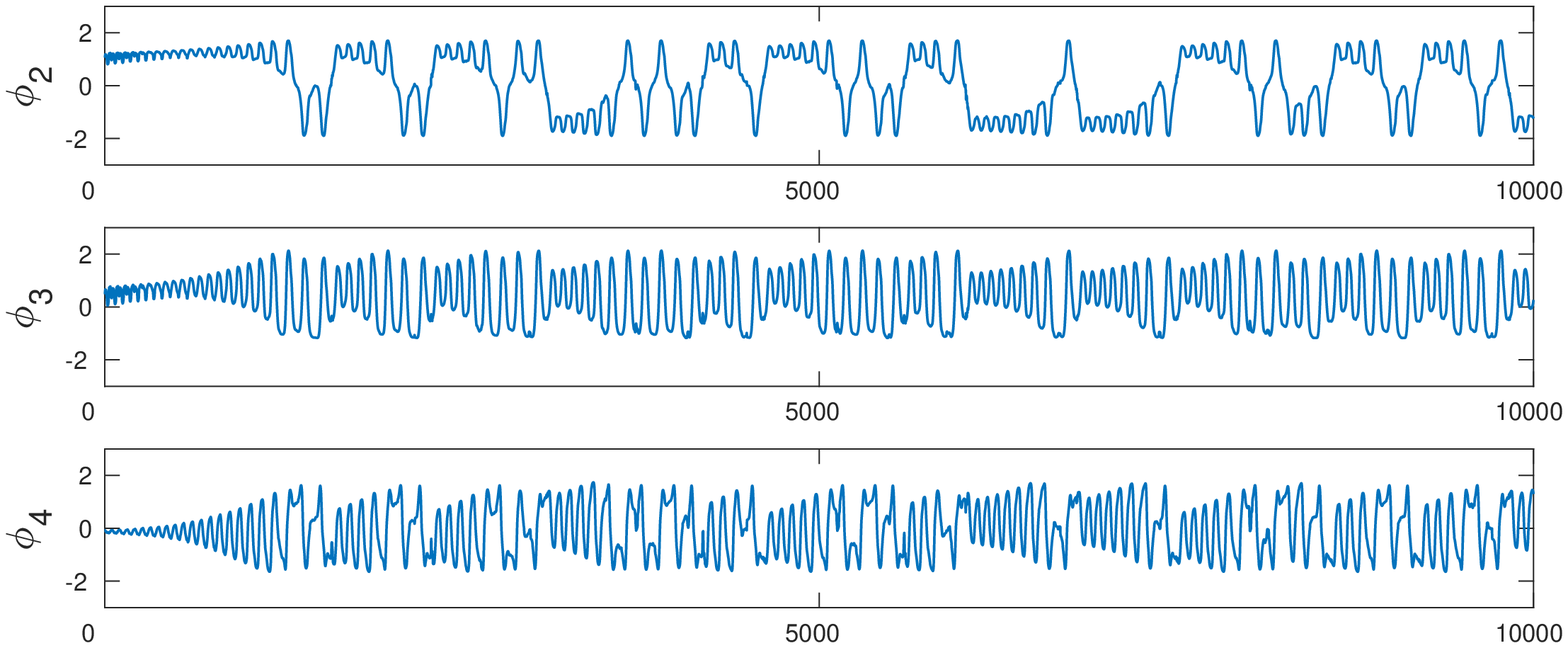}
	\includegraphics[trim = 0cm 0cm 2cm 0cm, clip, width=0.9\textwidth]{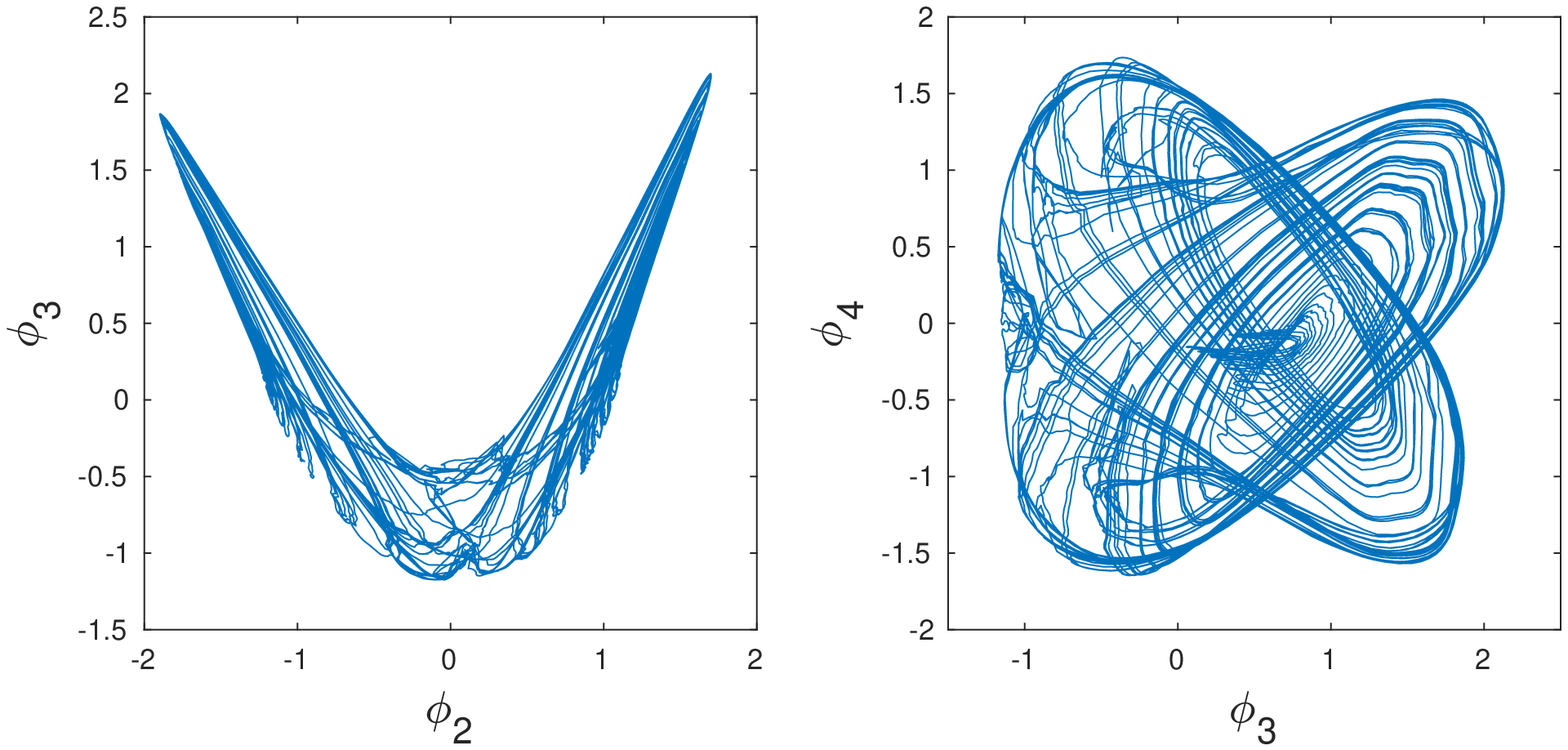}
	\caption{\textbf{Top:} Time series of the eigenvectors of the Lorenz 63 system \eqref{eqn:l63} partially observed through $f=(f^1,f^2)$ from 	\eqref{eqn:obs_l63}, used in the spatial reconstructions in Fig.\ \ref{fig:lorenz_x1x2_phis} for the first 10,000 samples. The first eigenfunction is the constant vector of ones. \textbf{Bottom:} Examples of two-dimensional representations of the eigenvectors.
	\label{fig:lorenz_x1x2_phis1D_2D}}
\end{figure}

\subsection{Moment reconstruction for the realtime multivariate MJO (RMM) index}
\label{subsect:RMM}

The phenomenon from climate science that we study here is known as the Madden-Julian oscillation (MJO; \cite{MaddenJulian71, MaddenJulian72}). MJO is the main tropical intraseasonal oscillation (ISO), and it corresponds to a 30-90-day eastward-propagating wave pattern with zonal wavenumber 1-4.

\paragraph{The observation map}  Among the multitude of indices that measure the MJO, the most common is the realtime multivariate Madden-Jullian oscillation (RMM) index \citep{WheelerHendon04}. RMM is a combined measure of the first two empirical orthogonal functions (EOFs) -- or principal components (PCs) -- of bandpass-filtered, and equatorially averaged outgoing longwave radiation, and 200hPa and 850hPa zonal wind data. In this experiment we use our framework to extract temporal patterns from the RMM index, and then show that the first four moments (mean, standard deviation, skewness and kurtosis) of the PDFs can be accurately reconstructed using only a small number of the leading eigenvectors. 

The dataset covers 23 years from September 1983 to June 2006, sampled once a day $\delta t=1$, that is, a total of $N=8337$ samples. We set the number of nearest neighbours to $k=100$, and $\epsilon=0.02$ as the width of the Gaussian kernel. We choose the embedding window to be $\Delta t=60$ days ($R=60$ timesteps) as it represents the average time of an MJO (30-90 days).  

Figure \ref{fig:corrErrorRMM} (Left) shows the absolute values of the expansion coefficients $\bm{c}_l$ of the $M=50$ leading eigenvectors computed using (\ref{eq:expansionCoef}) for the first four moments of the distributions -- the sign of the coefficients depends on the sign of the eigenvectors. We estimate the reconstruction error in Fig.\ \ref{fig:corrErrorRMM} (Right) for the leading $M=50$ eigenvectors, using Root Mean Square Error (RMSE) between the observed and the reconstructed time series: 

\[ \mbox{RMSE} = \sqrt{\frac{ \Vert \hat{\mathbb{E}}_n - \mathbb{E}_n \Vert_2^2}{N}}, \]

\noindent where $\mathbb{E}_n$ is the observed $n$-th moment of the distributions, and $\hat{\mathbb{E}}_n$ is the reconstructed $n$-th moment using only the leading few eigenvectors as in (\ref{eq:reconstrEigBasis}). Examples of reconstructions of the first four moments using $M \in \{5, 15, 50\}$ eigenvectors are shown in Figs.\ \ref{fig:RMM_moments_reconstructed5}, \ref{fig:RMM_moments_reconstructed15}, \ref{fig:RMM_moments_reconstructed50}, respectively. The optimal value of the number of eigenvectors needed to reconstruct each moment can be chosen using the decay of the error in Fig.\ \ref{fig:corrErrorRMM}. Thus, the mean and standard deviation are faithfully reconstructed using a very small number of eigenvectors (here approximately $M = 5$), while the skewness and kurtosis need a slightly higher number of eigenvectors (approximately $M = 15$ and $M = 50$, respectively). The correlation between the first non-constant eigenfunction $\bm{\phi}_2$ and the mean of RMM is 0.9611 (both have been normalized). Thus, some of the eigenvectors detected using our framework recover intrinsic properties of the statistical manifold, i.e., here $\bm{\phi}_2$ recovers the mean of the PDFs on the manifold. We also found that some of the eigenvectors are strongly correlated ($\approx 0.6-0.7$) with the standard deviation of the RMM index. Having the mean and the standard deviation could be very useful for example for prediction and uncertainty quantification when the past trajectory of the dynamical system is known. We tested our algorithm for robustness using the following parameter values: $\Delta t \in [30,90]$, $k \in [50,1000]$, $\epsilon \in [0.005, 2]$, and the results are very robust within these ranges. 

\begin{figure}[!ht]
	\center \includegraphics[trim = 0cm 0cm 0cm 0cm, clip, width=0.95\textwidth]{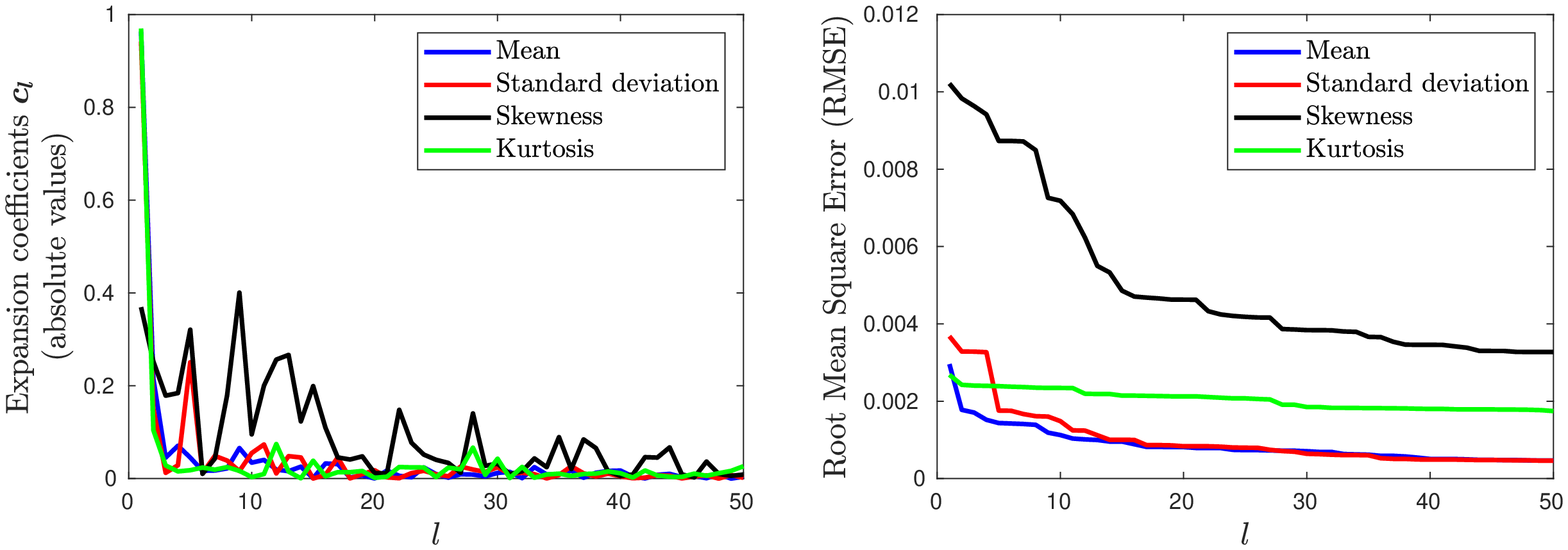}
	\caption{Reconstructions of the first four moments (mean, standard deviation, skewness and kurtosis) for the Real-time Multivariate MJO (RMM) index. (Left) Absolute values of the leading $M=50$ expansion coefficients $\bm{c}_l,\ 1 \leq l \leq M$. (Right) Root Mean Square Error (RMSE) using the leading eigenvectors. The error is computed between the observed and the reconstructed moments of RMM. All moments have been normalized to Euclidean norm 1. The eigenvectors form an orthonormal basis with respect to the inner product and the measure $\omega$. Examples of reconstructions using the leading $M=\{5, 15, 50\}$ eigenvectors are shown in Figs. \ref{fig:RMM_moments_reconstructed5}, \ref{fig:RMM_moments_reconstructed15} and \ref{fig:RMM_moments_reconstructed50}, respectively.
	\label{fig:corrErrorRMM}}
\end{figure}

\begin{figure}[!ht]
	\center \includegraphics[width=0.9\textwidth]{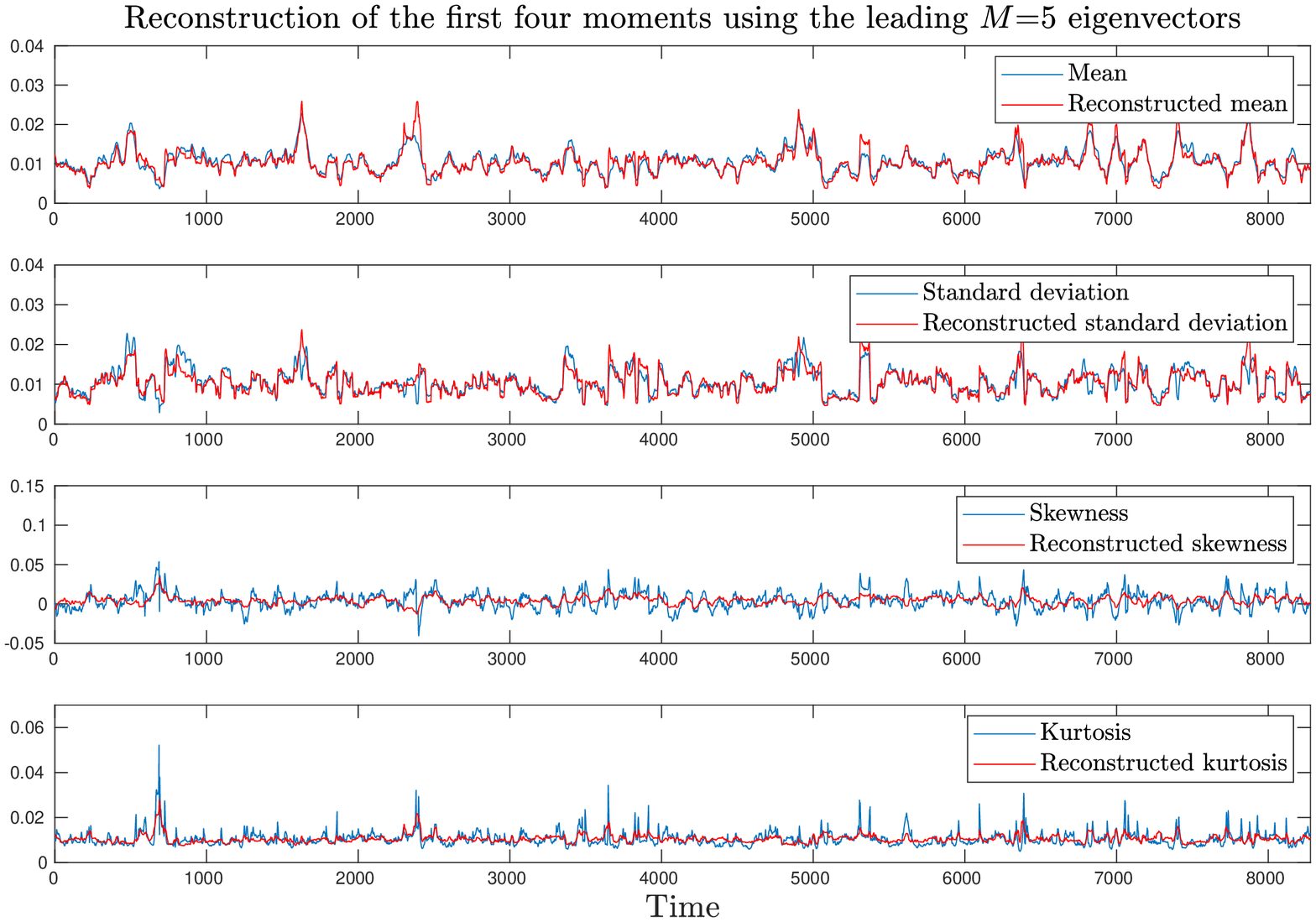}
	\caption{Reconstruction of the first four moments (mean, standard deviation, skewness and kurtosis) of RMM using the leading $M=5$ eigenvectors. The moments have been normalized to Euclidean norm 1. 
	\label{fig:RMM_moments_reconstructed5}}
\end{figure}

\begin{figure}[!ht]
	\center \includegraphics[width=0.9\textwidth]{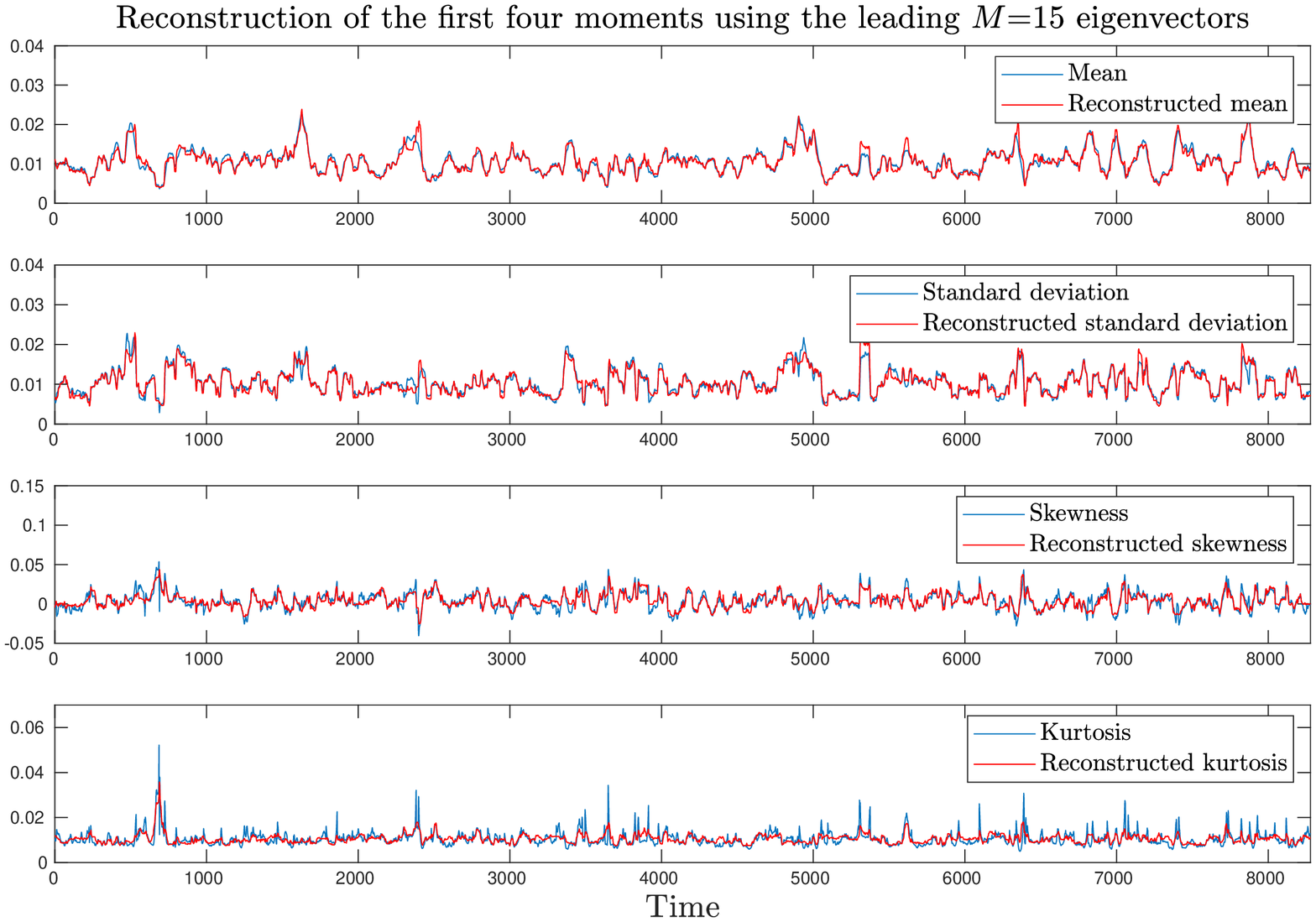}
	\caption{Reconstruction of the first four moments (mean, standard deviation, skewness and kurtosis) of RMM using the leading $M=15$ eigenvectors. The moments have been normalized to Euclidean norm 1.
	\label{fig:RMM_moments_reconstructed15}}
\end{figure}

\begin{figure}[!ht]
	\center \includegraphics[width=0.9\textwidth]{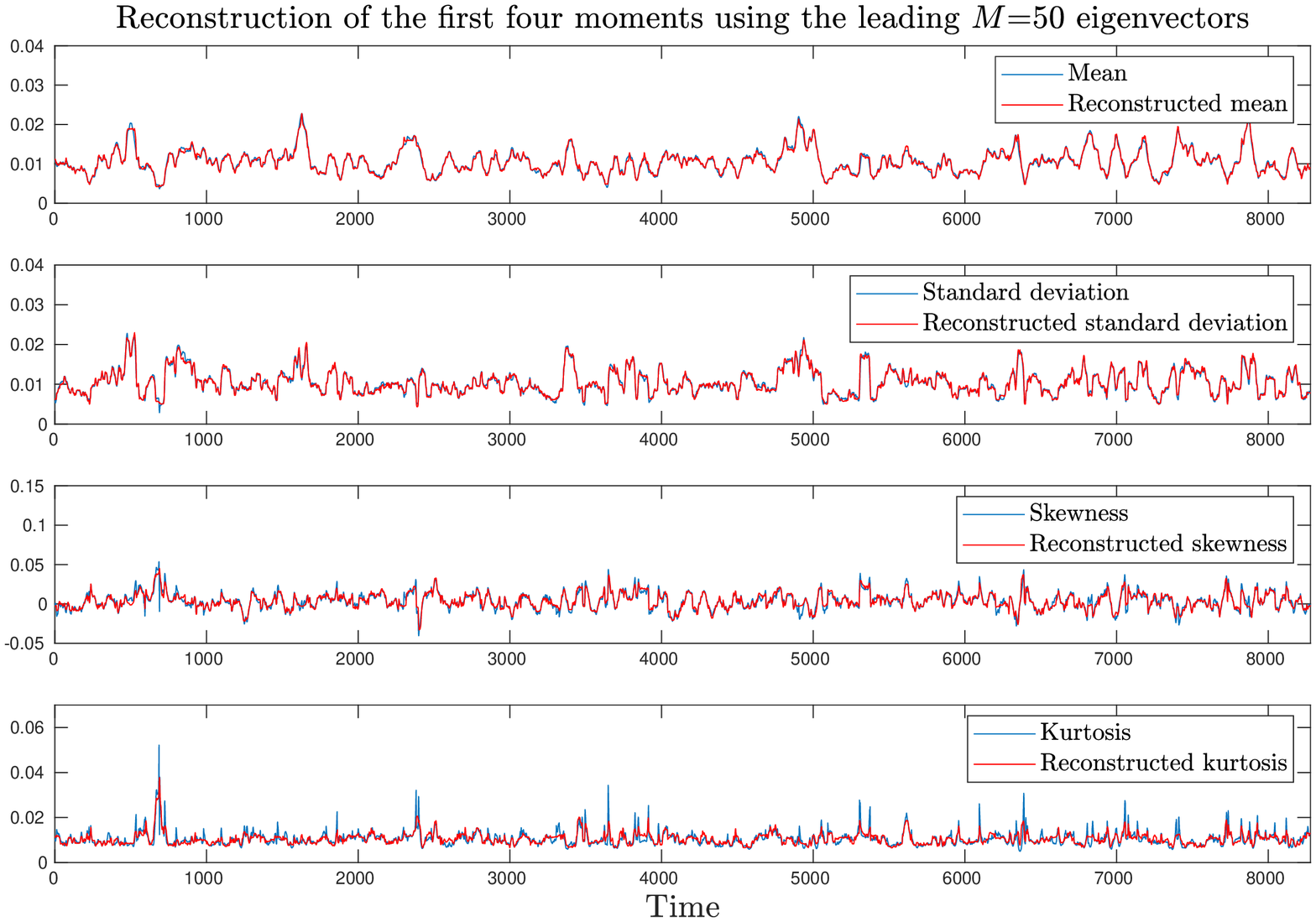}
	\caption{Reconstruction of the first four moments (mean, standard deviation, skewness and kurtosis) of RMM using the leading $M=50$ eigenvectors. The moments have been normalized to Euclidean norm 1.
	\label{fig:RMM_moments_reconstructed50}}
\end{figure}

\section{Conclusion}\label{sect:conclusion}

In this paper, we introduced a novel framework for feature extraction and moment reconstruction in dynamical systems that integrates ideas from machine learning, dynamical systems theory, and information geometry. Through dimension reduction, we extract temporal and spatiotemporal patterns of interest that describe the evolution of the dynamical system. While more conventional approaches for dimension reduction act directly in the original data spaces, our method acts on probability spaces. We use divergences between probability distributions on finite-time trajectories of the dynamical system, and thus are able to capture the dynamical evolution of the system. The divergences in the probability space allows us to define a kernel integral operator whose orthonormal eigenfunctions capture different timescales of the dynamical system (with emphasis on the slow timescales). If the collection of probability measures is in addition a manifold, we can equip this statistical manifold with the canonical Riemannian metric, allowing us to make connections to the field of information geometry.

One of our main results shows  that linear transformations on general functions over observables of the dynamical system can be written in terms of a time-averaging operator. For particular choices of these functions, these time-averaging operators are used to compute the moments of the collection of probability measures. We next exploited the fact that these transformations are linear combinations of the eigenfunctions of a kernel integral operator, and we showed that we can expand the moments of the distributions in this eigenfunction basis. This property provides a powerful tool that allows us to use nonparametric forecasting techniques based on out-of-sample extensions to predict the moments of the distributions at times further down the trajectory. We applied these techniques to three toy examples and a real-world atmosphere-ocean time series of the Madden-Julian oscillation, and showed that the (first four) moments of the distributions are reconstructed faithfully using only a few leading eigenfunctions.

The framework presented here opens up multiple possibilities for future work. For example, complex phenomena where data is generated by multiple heterogeneous sources, i.e., different ambient data spaces with different units, pose challenging issues for problems such as manifold alignment or multimodal data fusion and integration. Being intrinsically homogeneous, techniques that act directly on probability spaces allow for a coherent analysis of heterogeneous data. In this paper, the probability measures are estimated through kernel density estimation techniques, however this becomes intractable for high-dimensional observables (we worked here with one and two-dimensional observables). Kernel mean embedding of distributions \citep{SmolaEtAl07, MuandetEtAl17} are nonparametric techniques that map probability distributions into a reproducing kernel Hilbert space without requiring an explicit estimation of the probability distributions. Integration of these techniques with our framework would be very useful when dealing with high-dimensional observables. We work here with ergodic deterministic dynamical systems, but the framework could also be extended to stochastic dynamical systems \citep{BerryEtAl15}. In terms of applications, the framework presented here can be used to extract temporal patterns and forecast moments of the probability distributions of any dynamical system. In the future, we plan to apply the method to study other observables of the climate atmosphere ocean system, such as state-of-the-art El Ni\~no indices.

\section*{Acknowledgments}

S. Das and D. Giannakis gratefully acknowledge support from NSF Grant DMS 1854383, ONR Grant N00014-14-0150, ONR YIP Grant N00014-16-1-2649 and ONR MURI grant N00014-19-1-2421. E. Székely acknowledges support from ONR MURI grant 25-74200-F7112 and Grant/Project MM/SERP/CNRS/2013/ INT-10/002 from the Ministry of Earth Sciences, Government of India while a postdoctoral researcher at New York University. The authors would like to thank Jane Zhao for fruitful discussions. 

\begin{appendices}

\section{Choice of parameters}\label{app:distanceDistrib}

In Fig.\ \ref{fig:distdistrib} we display the distance distributions of the first $k$ nearest neighbors for the four models in Fig.\ \ref{fig:data} for different values of $k$, and we discuss an empirical way of choosing the optimal number of nearest neighbors in order to obtain good embeddings. As the number of nearest neighbors $k$ increases, the distance distributions shift from a positive skewness towards a negative skewness. In our experiments we observed that the best results were obtained when minimizing the skewness of the distance distributions of the nearest neighbors (here we used a symmetrized distance matrix). This is in part explained by the fact that a large positive skewness corresponds to a neighborhood graph where only very small neighborhoods are connected, while a large negative skewness indicates that there are a lot of edges connecting far away neighborhoods, thus leading to dense graphs. On the other hand, the neutral (no) skewness gives equal weight to both small and large distances creating a balanced neighborhood graph. 

For our experiments, we chose the values of $k$ that minimize the skewness, that is $k=500$ for Model I on the 2-torus, $k=7,000$ for Model II on the 2-torus, $k=3,000$ for the fixed point torus, and $k=2,000$ for the Lorenz system. For all models in Fig.\ \ref{fig:data}, the results were robust for ranges of values that guaranteed a small skewness. Thus, for Model I embeddings using a number of nearest neighbors in the range $k=[500,1000]$ were performing the best, as were embeddings using $k=[5000,10000]$ for Model II. 

Concerning the width of the Gaussian kernel, its choice is tightly related to the mean of the distance distribution to the $k$ nearest neighbors, which is around $\varepsilon=0.5$, with the bandwidth in \eqref{eq:def:Gauss} $\epsilon = 2 \varepsilon^2$. In the numerical experiments, we show results using $\varepsilon$ that provides the best visual embeddings, but our results have shown to be robust for values in the range $\varepsilon = [0.2, 1]$. Kernel similarities for five random data points for each of the four models are shown in Fig.\ \ref{fig:kernelSim}. We see for example that the kernel decays significantly faster for Model I compared to Model II on the 2-torus, thus directly influencing the choice of the optimal parameter values (both $k$ and $\epsilon$). By truncating the kernel at $k$ nearest neighbors, we remove the smoothness, however previous results \citep{TingEtAl10} have shown convergence of the graph Laplacian even for non-smooth kernels. The Hellinger distance being upper bounded by 1, the kernel similarities will be lower bounded by $e^{-\frac{1}{2\varepsilon^2}}$.

In this paper, the parameters $k$ and $\epsilon$ are global to all data points, but as the densities and the neighborhoods on the manifold change (e.g., for Model I the flow along the manifold speeds up or slows down in different regions leading to different local behaviors), one solution would be to use an adaptive nearest neighbor and a variable-bandwidth kernel \citep{TingEtAl10}. We believe adaptive algorithms would improve our results as for example we see that the decay of the kernel similarities behaves differently for different data points (Fig.\ \ref{fig:kernelSim}). We leave this analysis for future work.

\begin{figure}
\subfigure[2-torus (Model I)]{\includegraphics[trim = 0cm 0cm 0cm 0cm, clip, width=0.5\textwidth]{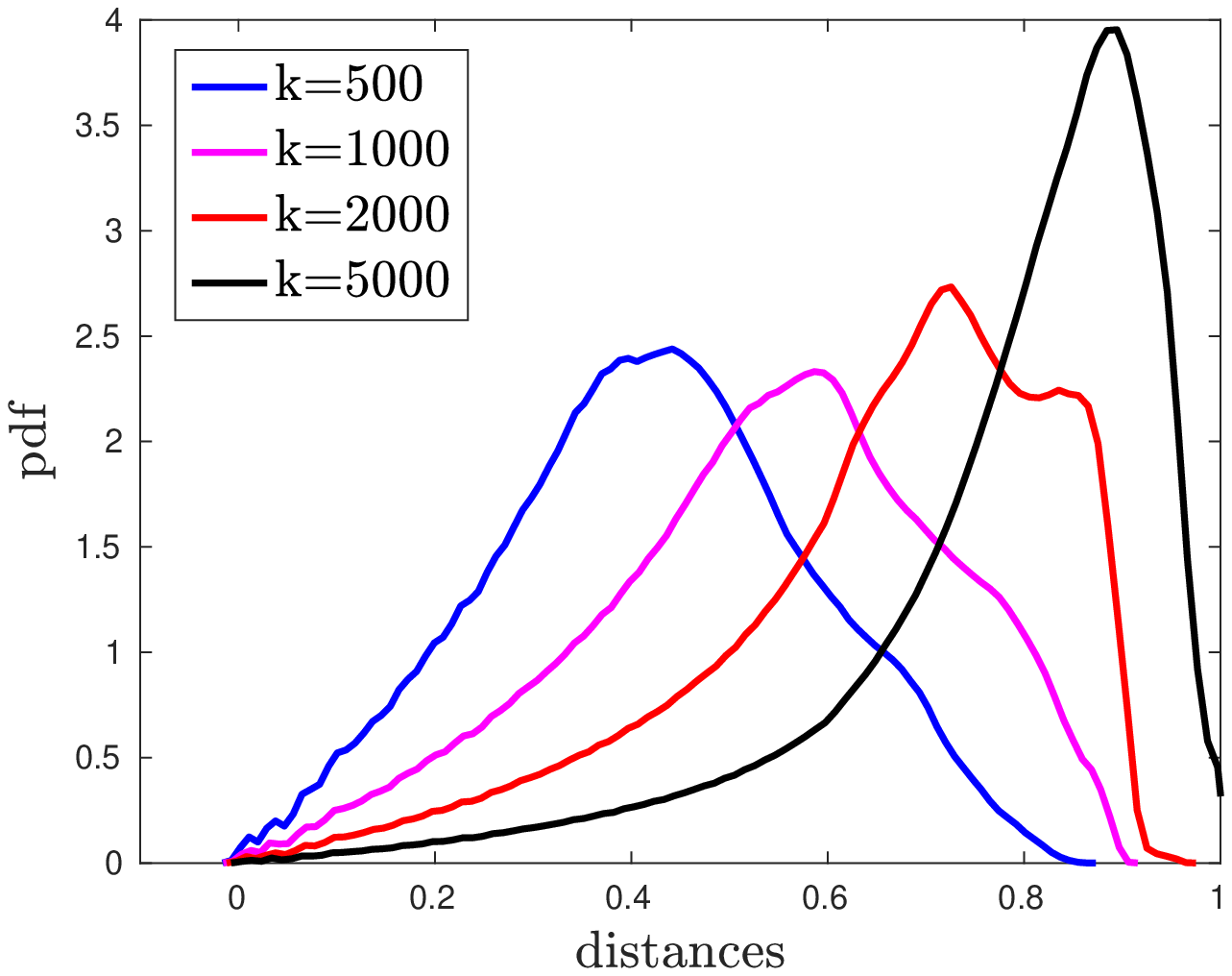}}
\subfigure[2-torus (Model II)]{\includegraphics[trim = 0cm 0cm 0cm 0cm, clip, width=0.5\textwidth]{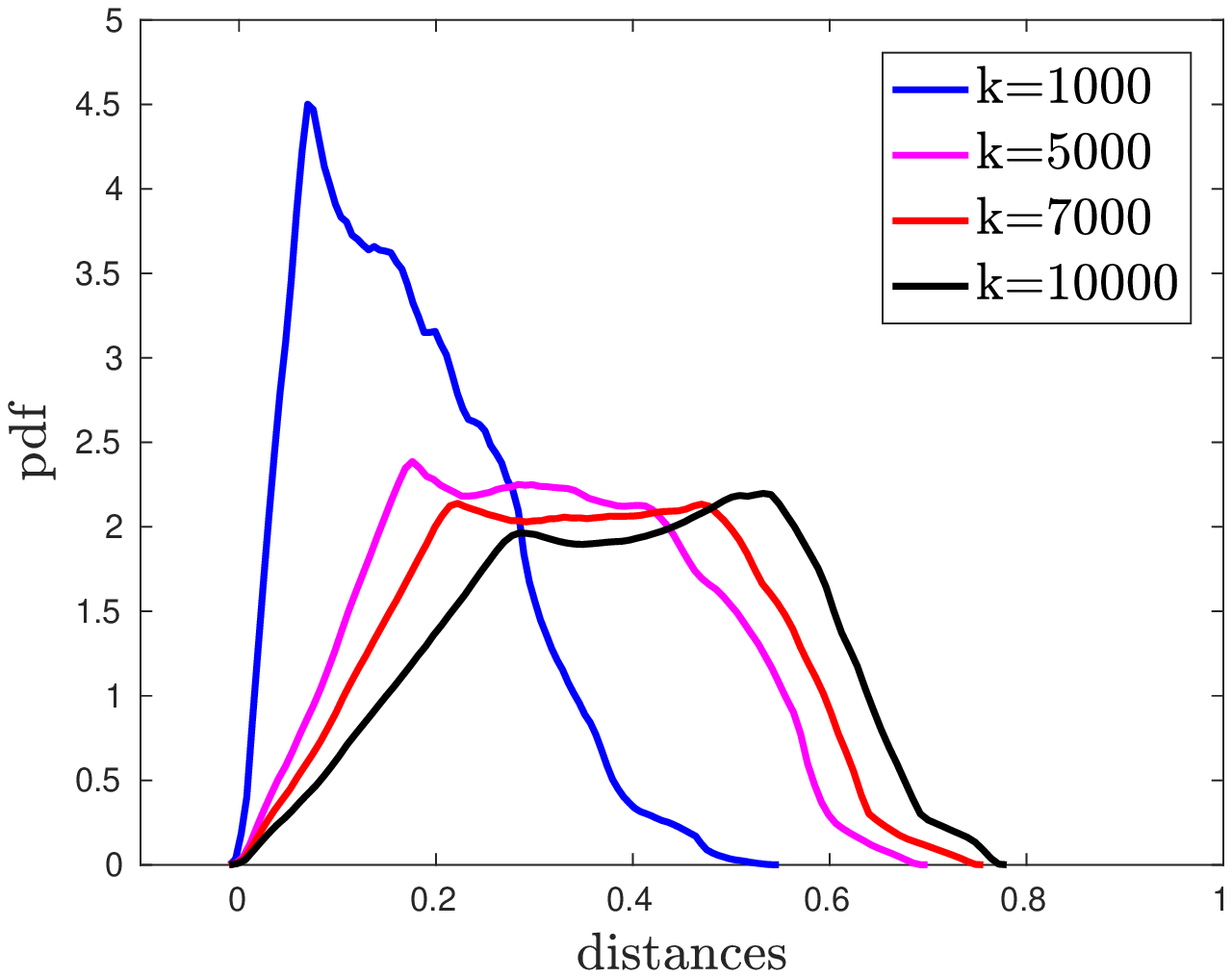}}
\subfigure[Fixed point torus]{\includegraphics[trim = 0cm 0cm 0cm 0cm, clip, width=0.5\textwidth]{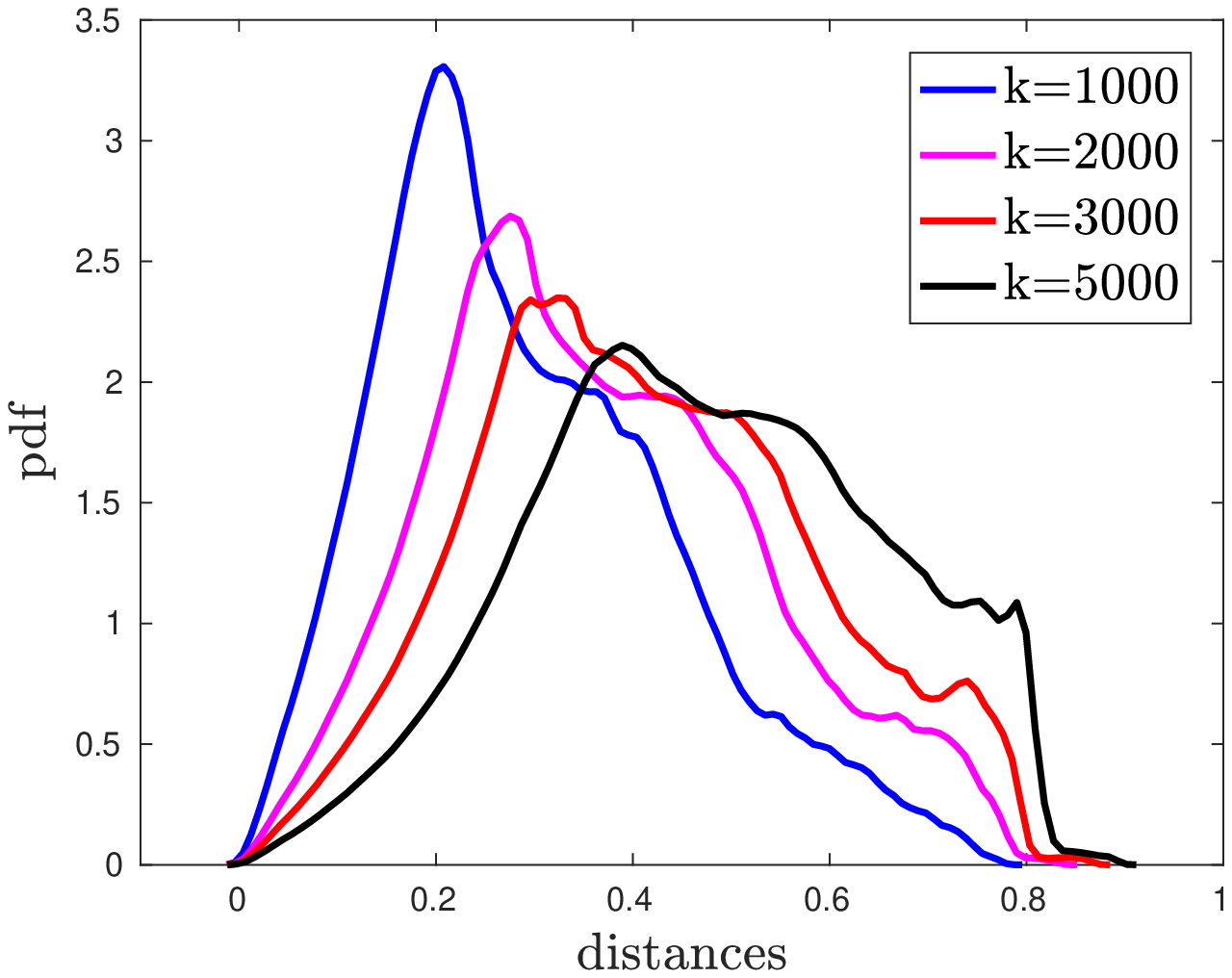}}
\subfigure[Lorenz 63]{\includegraphics[trim = 0cm 0cm 0cm 0cm, clip, width=0.5\textwidth]{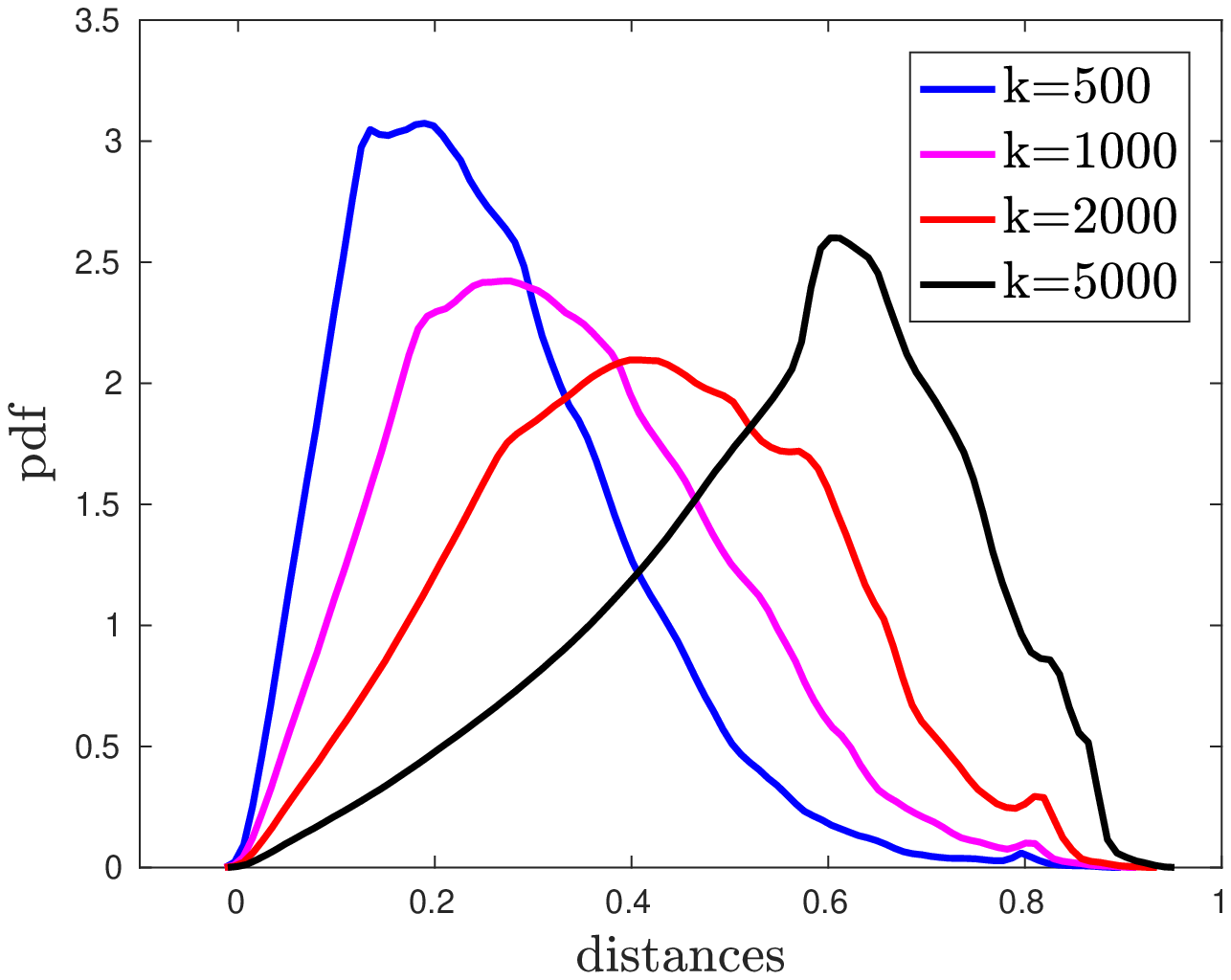}}
\caption{Distance distributions for different values of nearest neighbors for the four models used in the paper. Since the Hellinger distance is upper bounded by 1, the distance distributions shift from a positive skewness towards a negative skewness with the increase in $k$. When constructing the neighborhood graph we chose as the optimal $k$ the number of neighbors that reduced the skewness of the distributions.
\label{fig:distdistrib}}
\end{figure}

\begin{figure}
\subfigure[2-torus (Model I)]{\includegraphics[trim = 5.2cm 0cm 4.4cm 0cm, clip, width=\textwidth]{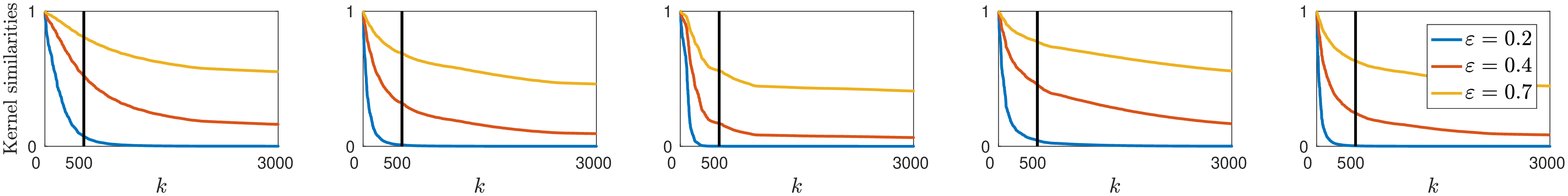}}
\subfigure[2-torus (Model II)]{\includegraphics[trim = 5.2cm 0cm 4.4cm 0cm, clip, width=\textwidth]{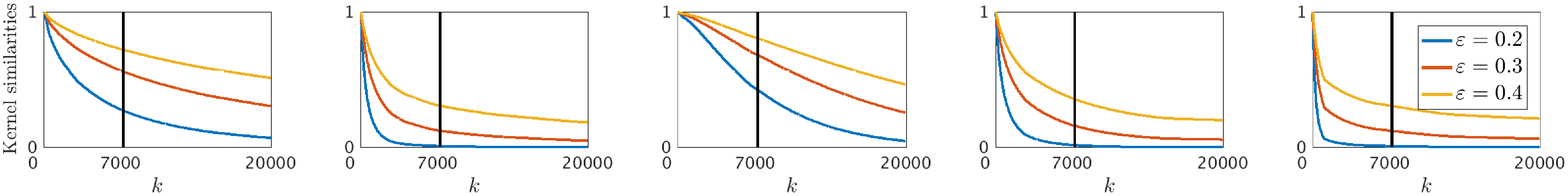}}
\subfigure[Fixed point torus]{\includegraphics[trim = 5.2cm 0cm 4.4cm 0cm, clip, width=\textwidth]{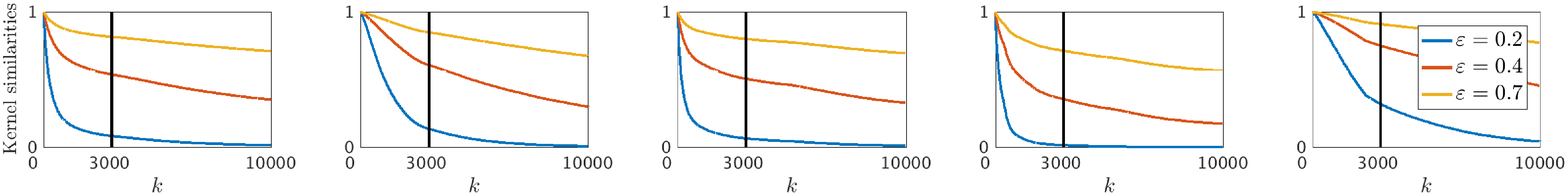}}
\subfigure[Lorenz 63]{\includegraphics[trim = 5.2cm 0cm 4.4cm 0cm, clip, width=\textwidth]{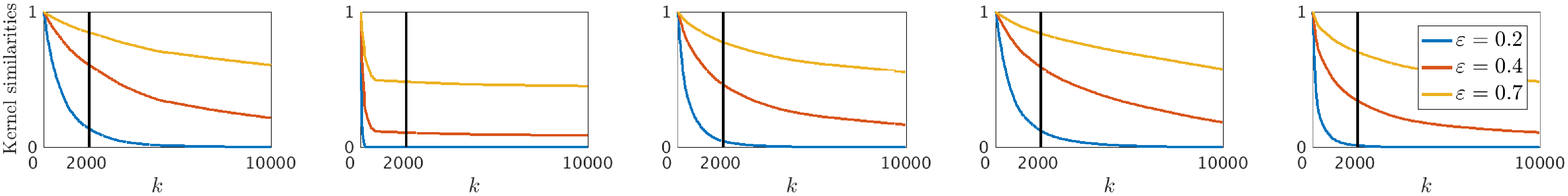}}
\caption{Kernel similarities for the four models in Fig.\ \ref{fig:data}, showing the exponential decay of the kernel  as a function of the number of nearest neighbors $k$. The vertical lines indicate the value of $k$ that we used in the paper for each of the numerical experiments, i.e., $k=\{500, 7000, 3000, 2000\}$. Since the Hellinger distance is upper bounded by 1, the kernel similarity will be lower bounded by $e^{-\frac{1}{2\varepsilon^2}}$.
\label{fig:kernelSim}}
\end{figure}

\end{appendices}

\bibliography{bibliography}
\end{document}